\documentclass{article}
\usepackage{sty/ijcai23}

\usepackage{times}
\usepackage{soul}
\usepackage{url}
\usepackage[hidelinks]{hyperref}
\usepackage[utf8]{inputenc}
\usepackage[small]{caption}
\usepackage[switch]{lineno}
\usepackage{graphicx}
\usepackage{amsmath}
\usepackage{amsthm}
\usepackage{booktabs}
\usepackage{algorithm}
\usepackage[noEnd=true,indLines=true,commentColor=black]{sty/algpseudocodex}
\usepackage{amsmath,amssymb,amsthm}
\usepackage{cleveref}
\usepackage{stmaryrd}
\usepackage[binary-units]{siunitx}
\usepackage[export]{adjustbox}
\usepackage{mfirstuc}
\usepackage{bm}
\usepackage{sty/mystyle}

\title{Quick Multi-Robot Motion Planning by Combining Sampling and Search}
\author{
Keisuke Okumura$^{1,2}$%
\footnote{
Most work done when Keisuke Okumura was a Ph.D. student at Tokyo Institute of Technology.
}
\and
Xavier D\'{e}fago$^3$
\affiliations
$^1$National Institute of Advanced Industrial Science and Technology (AIST)\\
$^2$University of Cambridge\\
$^3$Tokyo Institute of Technology
\emails
ko393@cl.cam.ac.uk, defago@c.titech.ac.jp
}

\begin{document}
\maketitle
\begin{abstract}
  We propose a novel algorithm to solve multi-robot motion planning (MRMP) rapidly, called \emph{Simultaneous Sampling-and-Search Planning (SSSP)}.
  Conventional MRMP studies mostly take the form of two-phase planning that constructs roadmaps and then finds inter-robot collision-free paths on those roadmaps.
  In contrast, SSSP simultaneously performs roadmap construction and collision-free pathfinding.
  This is realized by uniting techniques of single-robot sampling-based motion planning and search techniques of multi-agent pathfinding on discretized spaces.
  Doing so builds the small search space, leading to quick MRMP.
  SSSP ensures finding a solution eventually if exists.
  Our empirical evaluations in various scenarios demonstrate that SSSP significantly outperforms standard approaches to MRMP, i.e., solving more problem instances much faster.
  We also applied SSSP to planning for $32$ ground robots in a dense situation.
\end{abstract}

\section{Introduction}
Solving a \emph{multi-robot motion planning (MRMP)} problem within a realistic timeframe plays a crucial role in the modern and coming automation era, including fleet operations in warehouses~\cite{wurman2008coordinating} as well as collaborative robotic manipulation~\cite{feng2020overview}.
Nevertheless, the problem is known to be tremendously challenging even with simple settings~\cite{spirakis1984np,hopcroft1984complexity,hearn2005pspace}.
Filling this gap is a key milestone in automation.

Informally, MRMP aims at finding a collection of paths for multiple robots.
Those paths must be obstacle-free, and also, inter-robot collision-free.
Single-robot motion planning itself is intractable in general~\cite{reif1979complexity}, furthermore, the specific difficulty of MRMP comes from the second condition.
As a result, most studies on MRMP decouple the problem into \emph{(i)}~how to find obstacle-free paths and \emph{(ii)}~how to manage inter-robot collisions on those paths.
This decoupling takes the form of \emph{two-phase planning} that first constructs roadmaps and then performs \emph{multi-agent search}, finding collision-free paths on those roadmaps.
Here, a \emph{roadmap} is a graph that approximates the workspace for one robot and is carefully constructed to include obstacle-free paths from initial to goal states.
We can see two-phase planning examples in \cite{vsvestka1998coordinated,solovey2016finding,honig2018trajectory,solis2021representation}, to name just a few.

\paragraph{Contribution.}
In contrast to approaches based on two-phase planning, we propose a novel MRMP algorithm that \emph{simultaneously performs roadmap construction and multi-agent search}.
The crux is developing robot-wise roadmaps as necessary according to the multi-agent search progress.
Doing so keeps the search space small, leading to quick MRMP solving.
The proposed algorithm, called \emph{Simultaneous Sampling-and-Search Planning (SSSP)}, guarantees to eventually find a solution for solvable instances.
Although SSSP outputs \emph{sequential} solutions such that at most one robot moves at a time, it is possible to post-process known solutions to remove redundant motion and waiting time, as presented in this paper.
Our extensive evaluations on various scenarios with diverse degrees of freedom and kinematic constraints demonstrate that SSSP significantly outperforms standard approaches to MRMP, i.e., solving more problem instances much faster.
We also provide a planning demo with 32 ground robots (64-DOFs in total) in a dense situation.

\paragraph{Related Work.}
The (single-robot) motion planning problem, a fundamental problem in robotics, aims at finding an obstacle-free path in cluttered environments.
An established approach is \emph{sampling-based motion planning (SBMP)}~\cite{elbanhawi2014sampling}.
SBMP iteratively and randomly samples state points from the space and then constructs a roadmap;
a solution is derived by pathfinding on the roadmap.
Numerous SBMP algorithms have been developed so far, such as~\cite{kavraki1996probabilistic,lavalle1998rapidly,karaman2011sampling}.
SBMP can solve problems even with many degrees of freedom and has achieved successful results not limited to robotics~\cite{lavalle2006planning}.

In principle, SBMP is applicable to MRMP by considering one composite robot consisting of all robots~\cite{choset2005principles}, as seen in~\cite{sanchez2002delaying,le2018cooperative}.
However, such strategies require sampling from the high-dimensional space linear to the number of robots, being a bottleneck even for SBMP~\cite{lavalle2006planning}.
Consequently, recent studies mostly take the aforementioned two-phase planning.
In those studies, as the first phase, roadmaps are explicitly prepared via conventional SBMP~\cite{vsvestka1998coordinated,wagner2012probabilistic,solovey2016finding,solis2021representation,dayan2021near} or implicitly embedded as lattice grids~\cite{han2018sear,honig2018trajectory,cohen2019optimal}.
Depending on the heterogeneity of robots, a roadmap is shared among robots, or, robot-wise roadmaps are constructed.
The lattice grids are available when the configuration space of each robot is not high-dimensional or state transitions of robots are restricted to a limited number; otherwise, the search space dramatically grows.
This study does not assume such limitations.
The second phase often uses \emph{multi-agent pathfinding (MAPF)} algorithms such as~\cite{sharon2015conflict,barer2014suboptimal,wagner2015subdimensional}, including prioritized planning~\cite{erdmann1987multiple,silver2005cooperative}, and sometimes discretized versions of SBMP~\cite{cap2013multi,solovey2016finding}.
MAPF~\cite{stern2019def} is a problem of finding a set of collision-free paths on a graph and collects extensive attention since the 2010s.

SSSP is directly inspired by two algorithms, respectively for SBMP and MAPF:
\emph{(i)~Expansive Space Trees (EST)}~\cite{hsu1997path} is an example of SBMP, which performs planning by constructing a query tree growing with random walks.
\emph{(ii)~\astar with operator decomposition}~\cite{standley2010finding} solves MAPF efficiently by decomposing successors in the search tree such that at most one agent takes an action, rather than all agents take actions simultaneously.
We realize rapid MRMP by combining these techniques.

\paragraph{Paper Organization.}
\Cref{sec:pre} describes the problem formulation and assumptions for MRMP.
\Cref{sec:algo} describes SSSP.
\Cref{sec:eval} presents the empirical results.
\Cref{sec:discussion} provides discussions.
Throughout the paper, we present a sub-optimal algorithm and focus on the decision problem because solving MRMP itself is challenging.
Optimal SSSP is discussed at the end.
The appendix, code, and video are available at \url{https://kei18.github.io/sssp/}.

\section{Preliminaries}
\label{sec:pre}
\subsection{Problem Definition of MRMP}

We consider a problem of motion planning for a team of $n$ robots $A = \{1, 2,\ldots,n\}$ in the 3D closed \emph{workspace} $\W \subset \mathbb{R}^3$.
Each robot $i$ is operated in its own \emph{configuration space} $\C_i \subset \mathbb{R}^{d_i^\C}$, where $d_i^\C \in \mathbb{N}_{>0}$.
A set of points occupied by robot $i$ at a configuration $q \in \C_i$ is denoted as $\R_i(q) \subset \W$.
The space \W may contain obstacles $\O \subset \W$.
A {free space} for robot $i$ is then $\cfree_i \defeq \{ q \in \C_i \mid \R_i(q) \cap \O = \emptyset \}$.
A \emph{trajectory} for robot $i$ is defined by a continuous mapping $\sigma_i: \mathbb{R}_{\geq 0} \mapsto \C_i$.

\begin{definition}
  An \emph{MRMP instance} is defined by a tuple $($\W, $A$, \C, \O, \R, $\Q\init$, $\Q\goal)$,
  where
  $\C \defeq (\C_i)^{i\in A}$,
  $\R \defeq (\R_i)^{i\in A}$,
  $\Q\init \defeq \left( q_i\init \in \C_i \right)^{i \in A}$,
  and $\Q\goal \defeq ( Q_i\goal \subseteq \C_i)^{i \in A}$.
  \label{def:instance}
\end{definition}

\begin{definition}
  Given an MRMP instance, the \emph{MRMP problem} is to find a tuple of $n$ trajectories $(\sigma_i)^{i \in A}$ (i.e., solution) and $t\sub{end} \in \mathbb{R}_{\geq 0}$, satisfying the following conditions:
  \begin{itemize}
  \item \emph{endpoint}: $\sigma_i(0) = q_i\init \land \sigma_i(t\sub{end}) \in Q_i\goal$
  \item \emph{obstacle-free}: $\sigma_i(\tau) \in \cfree_i,\; 0 \leq \tau \leq t\sub{end}$
  \item \emph{inter-robot collision-free}:
      $\R_i\bigl(\sigma_i(\tau)\bigr) \cap \R_j\bigl(\sigma_j(\tau)\bigr) = \emptyset, i \neq j \in A, 0 \leq \tau \leq t\sub{end}$
  \end{itemize}
  \label{def:mrmp}
\end{definition}

\subsection{Constraints of Robot Motions}
\Cref{def:mrmp} assumes that a robot can go in any direction in the configuration space unless it encounters obstacles.
We call it \emph{geometric MRMP}.
Meanwhile, robots are often subject to kinematic and dynamics constraints.
\emph{Kinematic constraints} restrict the local directions of motion available to a robot from a given configuration.
For instance, wheeled robots cannot translate sideways.
\emph{Dynamics constraints} are governed by the time derivatives, such as velocity and acceleration.
For instance, cars cannot stop instantly.
Kinematic and dynamics constraints are collectively called \emph{kinodynamic constraints}.
Motion planning under kinodynamic constraints is called \emph{kinodynamic planning}~\cite{donald1993kinodynamic}.
Since this study is an early-stage attempt at combining sampling and search, \emph{we consider only kinematic constraints and ignore dynamics constraints} for simplicity.

\subsection{Planning with Kinematic Constraints}
Consider a \emph{control space} $\U_i \in \mathbb{R}^{d_i^\U}$ for robot $i$, where $d_i^\U \in \mathbb{N}_{>0}$.
For instance, the control space of a wheeled robot is defined by motor controls of its wheels.
Then, transitions of configurations under kinematic constraints are governed by $\dot{q} = f_i(q, u)$, where $q \in \C_i$ and $u \in \U_i$.
A \emph{kinematic MRMP instance} is defined by a composition of an MRMP instance in \cref{def:instance}, control spaces $(\U_i)^{i\in A}$, and transition functions $(f_i)^{i\in A}$.
Given a kinematic MRMP instance, the \emph{kinematic MRMP problem} asks a sequence of control inputs for each robot $i$, that is, $\xi_i: \mathbb{R}_{\geq 0} \mapsto \U_i$.
A trajectory $\sigma_i$ of configurations is successively defined as $\sigma_i(0) \defeq q_i\init$ and $\sigma_i(t) \defeq \int_{0}^{t} f_i\left(\sigma_i(\tau), \xi_i(\tau)\right)d\tau + \sigma_i(0)$.
$\xi_i$ constitutes a \emph{solution} when $\sigma_i$ satisfies the conditions in \cref{def:mrmp} for all robots.

\subsection{Discretized Time and Local Planner}
Both geometric and kinematic MRMPs are defined in continuous time.
However, it is realistic for planning to discretize the time.
That is, by introducing $\dt \in \mathbb{R}_{>0}$ as a small amount of time, we aim at finding a path of configurations for each robot, such that any consecutive two points are travelable in \dt, without encountering obstacles, without inter-robot collisions, and following kinematic constraints.
For instance, given a sequence of control inputs $\xi^0, \xi^1, \ldots$, a path of configurations in kinematic MRMP is successively defined as $\sigma_i^0 \defeq q_i\init$ and $\sigma_i^k \defeq \sigma_i^{k-1} + f_i\bigl( \sigma_i^{k-1}, \xi^{k-1} \bigr)\dt$.
Herein, $\delta$ can be regarded as a problem input shared between robots.

For this discretization, as often assumed in motion planning studies~\cite{choset2005principles,lavalle2006planning}, we assume that each robot $i$ has a \emph{local planner} denoted as $\connect_i$.
Given two configurations $q\from, q\to \in \C_i$, this function returns a unique trajectory $\sigma$ of the duration \dt that satisfies:
\emph{(i)}~$\sigma(0) = q\from \land \sigma(\dt) = q\to$,
\emph{(ii)}~$\sigma(\tau) \in \cfree_i$ for $0 \leq \tau \leq \dt$,
and \emph{(iii)}~$\sigma$ follows $f_i$ of kinematic MRMP.
If no such $\sigma$ is found, $\connect_i$ returns $\bot$.
For instance, $\connect_i(q\from, q\to)$ may output $(\dt - \tau)q\from + \tau q\to$ for geometric MRMP, or Dubins paths~\cite{dubins1957curves} for car-like robots.
Since kinematic MRMP excludes dynamics constraints, this paper assumes that each robot can always remain in its current configuration, i.e., $\connect_i(q, q) \neq \bot$ for any $q \in \cfree_i$.

Observe that the local planners \emph{hide} control inputs of kinematic MRMP;
as long as they are definable, we can directly consider planning in configuration spaces.
Therefore, we collectively call geometric and kinematic MRMPs the \emph{MRMP problem} and do not distinguish the two explicitly.
With the local planners, a \emph{solution} of MRMP is a tuple of paths $(\Pi_i)^{i \in A}$, where $\Pi_i = (q_0, q_1, \ldots, q_k \mid q_t \in \C_i )$ and $k$ is common between robots, satisfying the following conditions:
\begin{itemize}
\item endpoint:~$\loc{i}{0} = q\init_i \land \loc{i}{k} \in \Q\goal_i$
\item consistent path:~$\connect_i(\loc{i}{t}, \loc{i}{t+1}) \neq \bot$
\item inter-robot collision-free:
\end{itemize}
\begin{align}
  \begin{split}
  &\R_i\bigl( \sigma_i(\tau) \bigr) \cap \R_j\bigl( \sigma_j(\tau) \bigr) = \emptyset,\;
  0 \leq \tau \leq \Delta, i \neq j \in A,
  \\
  &\sigma_{\{i, j\}} = \connect_{\{i, j\}} (\loc{\{i, j\}}{t}, \loc{\{i, j\}}{t+1})
  \end{split}
  \label{eq:collision}
\end{align}
The domain of $t$ is $\{ 0, 1, \ldots, k - 1 \}$.

\subsection{Roadmap}
A \emph{roadmap} for robot $i$ is a directed graph $G_i = (V_i, E_i)$ which approximates $\cfree_i$ with a finite set of vertices.
Each vertex $q \in V_i$ corresponds to a configuration of $\C_i$, thus simply denoted as $q \in \C_i$.
The roadmap $G_i$ must satisfy $q \in \cfree_i$ for all $q \in V_i$ and $\connect_i(q, q') \neq \bot$ for all $(q, q') \in E_i$.

\subsection{Blackbox Utility Functions}
To solve MRMP, we introduce four functions.
The first three are common in motion planning studies~\cite{choset2005principles,lavalle2006planning}, whereas the last one is specific to MRMP.

\paragraph{Sampling.}
The function $\sample_i$ randomly samples a configuration $q \in \C_i$, where $q$ may not be in $\cfree_i$.

\paragraph{Distance.}
The function $\dist_i: \C_i \times \C_i \mapsto \mathbb{R}_{\geq 0}$ defines a distance of two configurations of robot $i$.
It is not necessary for this function to consider obstacles or inter-robot collisions, however, we assume that $p = q \Leftrightarrow \dist_i(p, q) = 0$, $\dist(p, q) = \dist(q, p)$, and $\dist_i$ satisfies the triangle inequality, e.g., the Euclidean distance.

\paragraph{Steering.}
The function $\steer_i$ takes two configurations $q\from$, $q\to \in \cfree_i$ and then returns a ``closer'' configuration $q \in \cfree_i$ to $q\to$.
With a prespecified parameter $\epsilon > 0$, formally:
\begin{align*}
  \begin{split}
    &\steer_i\left(q\from, q\to\right) \defeq \argmin_{q \in U_\epsilon} \dist_i\left(q, q\to\right)
    \;\text{where} \\
    &U_\epsilon \defeq \left\{ q \in \cfree_i \mid \dist_i\left(q\from, q\right) \leq \epsilon,
    \connect_i\left(q\from, q\right) \neq \bot \right\}
  \end{split}
\end{align*}
In practice, \steer can be approximately implemented by binary search, e.g., starting from $q\to$, repeatedly sampling $q \in \C_i$ while halving the distance from $q\from$ until $q$ is feasible.

\paragraph{Collision.}
For two robots $i, j$, given four configurations $q_{\{i, j\}}\from$, $q_{\{i, j\}}\to \in \cfree_{\{i, j\}}$ such that $\connect_{\{i, j\}}(q_{\{i, j\}}\from, q_{\{i, j\}}\to) \neq \bot$, the function $\collide_{(i, j)}(q_i\from, q_i\to, q_j\from, q_j\to)$ returns \true when there is a collision if two robots simultaneously change their configurations from $q_{\{i, j\}}\from$ to $q_{\{i, j\}}\to$.
Here, a \emph{collision} is defined similarly to \cref{eq:collision}.
The function otherwise returns \false.
For convenience, we use $\collide(\Q\from, \Q\to)$, where $\Q\suf{\{from, to\}} = (q_i\suf{\{from, to\}} \in \C_i\suf{free})^{i \in A}$.
This shorthand notation returns \true if and only if there is a pair $(i, j)$ for which \\$\collide_{(i, j)}(q_i\from,q_i\to,q_j\from, q_j\to)$ returns \true.
As for implementation, since collision detection is important in single-robot motion planning and has been studied for a long time, well-known open-source libraries are available, e.g.,~\cite{pan2012fcl}.

\paragraph{Domain Independence.}
To sum up, we solve MRMP using only five blackbox functions: \connect, \sample, \dist, \steer, and \collide.
Doing so makes our approach non-restrictive to specific robotic systems, rather it is applicable to many planning domains as we will see in the experiments.

\section{Algorithm Description}
\label{sec:algo}

In a nutshell, SSSP performs a best-first search using operator decomposition~\cite{standley2010finding} while simultaneously growing robot-wise roadmaps via random walks~\cite{hsu1997path}.
This section first explains the core idea, followed by the pseudocode, theoretical analysis of completeness, and postprocessing to obtain better solutions.

\subsection{Core Idea}
As a high-level description, SSSP constructs a search tree while expanding robot-wise roadmaps.
Each search node in the tree contains a tuple of configurations $\Q = (q_i \in \cfree_i)^{i \in A}$ and robot $i$ that will take the next action.
When this node is selected during the search process, the algorithm does \emph{vertex expansion} and \emph{search node expansion} in order.
The former expands the roadmap $G_i$ for robot $i$ by random walks from $\Q[i]$.
The latter creates the successors of the node by transiting the configuration of robot $i$ from $\Q[i]$ to its neighboring configurations on $G_i$, and then passing the turn for $i$ to $i + 1$.
\Cref{fig:concept} illustrates this procedure with initial roadmap construction explained later, while \cref{fig:roadmap-example} shows an example of constructed roadmaps.

{
  \begin{figure}[t!]
    \centering
    \includegraphics[width=1.0\linewidth]{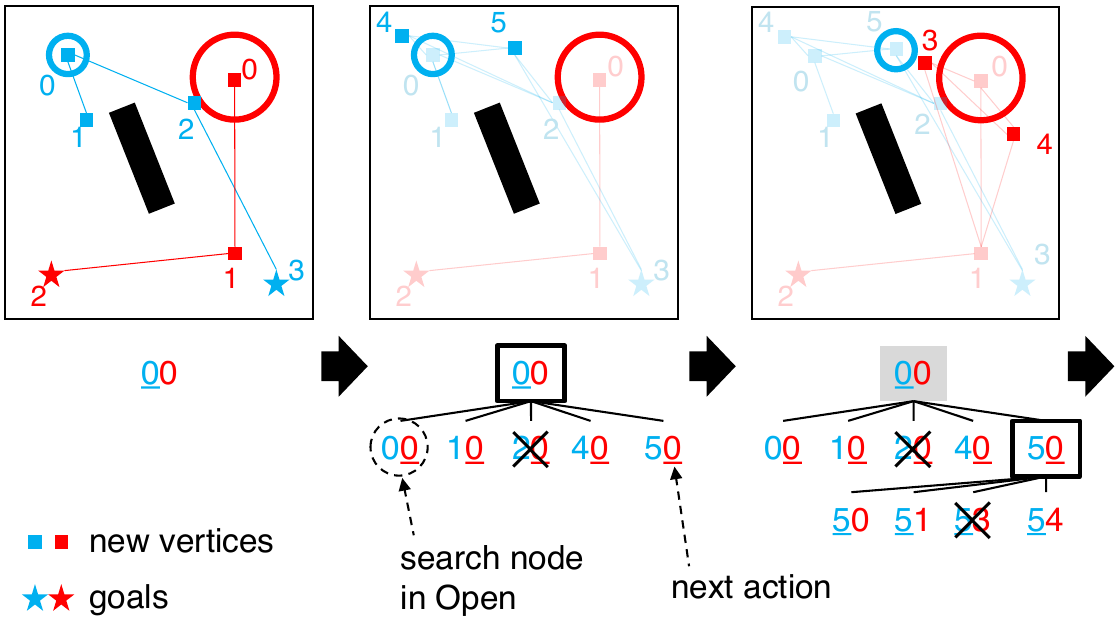}
    \caption{
      Illustration of SSSP.
      The algorithm progresses from left to right.
      \emph{top:}~Robot-wise roadmaps and search situations.
      Two robots are shown by colored circles and an obstacle by a black rectangle.
      \emph{bottom:}~Search trees.
      A node `\underline{x}y' corresponds to a situation where the blue and red robots are respectively at vertices `x' and `y,' and the blue robot will take the next action.
      \emph{left:}~Initial roadmaps.
      The search starts from the node `\underline{0}0.'
      \emph{middle:}~Vertex expansion and search node expansion for the blue robot.
      The node `2\underline{0}' is not generated due to inter-robot collision.
      \emph{right:}~The red robot's turn.
      The search continues until all robots reach their goals.
    }
    \label{fig:concept}
  \end{figure}
}

{
  \setlength{\tabcolsep}{0pt}
  \newcommand{\figcol}[1]{
    \begin{minipage}{0.242\linewidth}
      \centering
      \includegraphics[width=1.0\linewidth]{fig/raw/#1}
    \end{minipage}
  }

  \begin{figure}[t!]
    \centering
    \begin{tabular}{cccc}
      \figcol{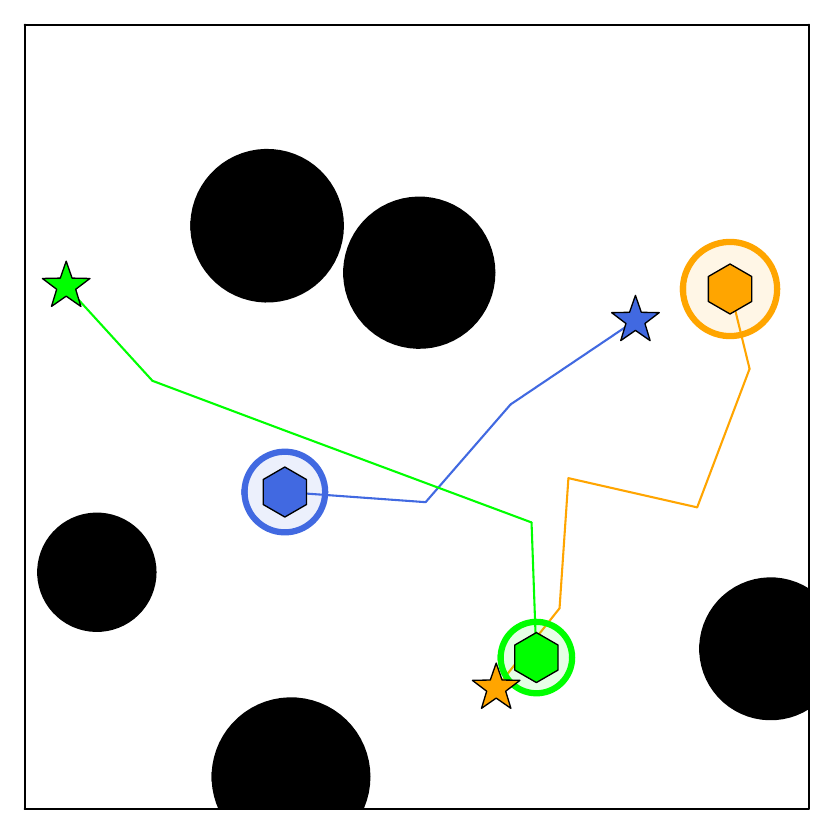}
      & \figcol{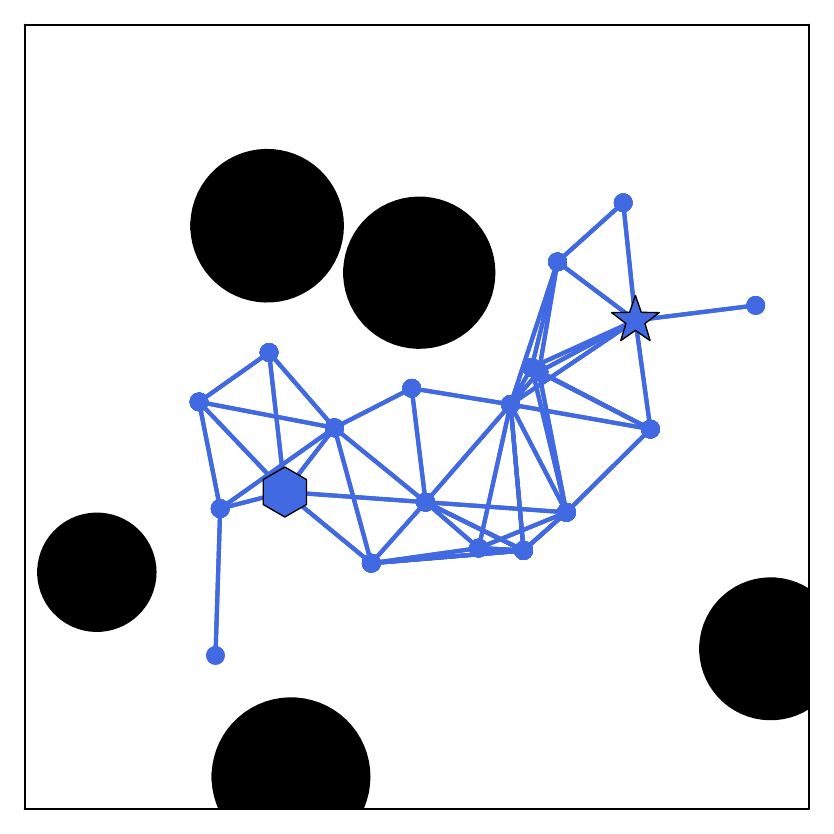}
      & \figcol{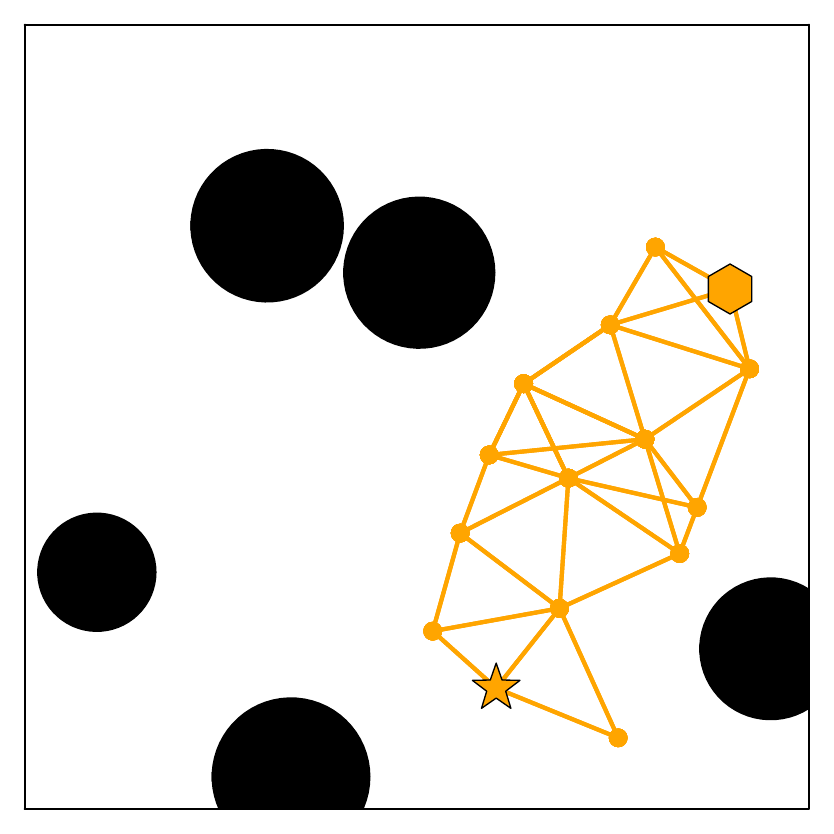}
      & \figcol{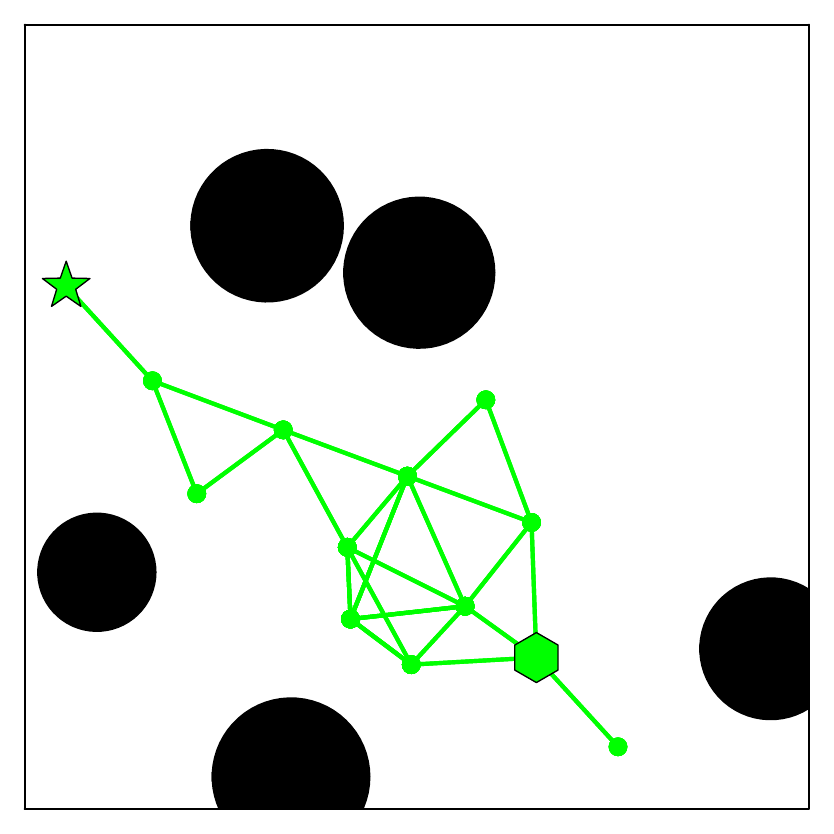}
    \end{tabular}
    \caption{
      Example of constructed roadmaps for 2D circular robots (i.e., \scenario{point2d} in \cref{fig:result}).
      A solution is depicted at the leftmost.
      The others show respective roadmaps for each robot.
    }
    \label{fig:roadmap-example}
  \end{figure}
}

{
  \newcommand{\Hyperparams}[1]{\AlgoPreface{params}{#1}}
  \newcommand{\append}{\funcname{append}}
  \begin{algorithm}[th!]
    \caption{SSSP; \textbf{input}:~instance $I$; \textbf{output}:~solution}
    \label{algo:planner}
    %
    \begin{algorithmic}[1]
      \Hyperparams{\#sampling $m \in \mathbb{N}_{>0}$}
      \item[]~~~~~~~~random sampling prob. $\lambda \in (0, 1{]}$
      \item[]~~~~~~~~threshold distances $\theta_i \in \mathbb{R}_{> 0}$, decay rate $\gamma \in (0, 1)$
      \State $G_i = (V_i, E_i) \leftarrow \funcname{init\_roadmap}(I, i)$; for each $i \in A$
      \label{algo:planner:init-roadmap}
      \While{\true}
      \label{algo:planner:search-iteration}
      \State initialize \open, \explored
      \label{algo:planner:init-open}
      \Comment{priority queue, set}
      \State $\open.\push\left(
      \left\langle
      \Q: \Q\init,\;
      \nextagent: 1,\;
      \parent: \bot
      \right\rangle
      \right)$
      \State $\explored.\append\left(\left(\Q\init, 1\right)\right)$
      \While{$\open \neq \emptyset$}
      \State $\N \leftarrow \open.\pop()$
      \label{algo:planner:pop}
      \If{$\forall i \in A, \N.\Q[i] \in \Q\goal_i$}
      \label{algo:planner:goal-check}
      \State \Return $\backtrack(\N)$
      \EndIf
      \State $i \leftarrow \N.\nextagent$; $q\from \leftarrow \N.\Q\from[i]$
      \For{$1, 2, \ldots, m$}
      \Comment{vertex expansion via sampling}
      \label{algo:planner:vertex-expansion}
      \State $q\new \leftarrow \sample_i()$
      \State with prob. $(1-\lambda)$: $q\new \leftarrow \steer_i(q\from, q\new)$
      \label{algo:planner:sampling}
      \If{$\min_{q \in V_i}\dist_i\left(q, q\new\right) > \theta_i$}
      \label{algo:planner:distance-check}
      \State $V_i \leftarrow V_i \cup \left\{ q\new \right\}$; update $E_i$
      \label{algo:planner:add-vertex}
      \EndIf
      \EndFor
      \label{algo:planner:vertex-expansion-end}
      \State $j \leftarrow i+1~\text{\textbf{if}}~i \neq |A|~\text{\textbf{else}}~1$
      \label{algo:planner:search-node-expansion}
      \Comment{node expansion}
      \For{$q\to \in \left\{ q \in V_i \mid \left(q\from, q\right) \in E_i \right\}$}
      \State $\Q\new \leftarrow \cpy(\N.\Q)$; $\Q\new[i] \leftarrow q\to$
      \IfSingle{$\left(\Q\new, j\right) \in \explored$}{\Continue}
      \IfSingle{$\funcname{collide}\left(N.\Q, Q\new\right)$}{\Continue}
      \label{algo:planner:collision-check}
      \State $\open.\push\left(
      \left\langle
      \Q: \Q\new,
      \nextagent: j,
      \parent: \N
      \right\rangle
      \right)$
      \State $\explored.\append\left((\Q\new, j)\right)$

      \EndFor
      \label{algo:planner:search-node-expansion-end}
      \EndWhile
      \label{algo:planner:end-search}
      \State $\theta_i \leftarrow \gamma\theta_i$; for each $i \in A$
      \label{algo:planner:decrease-theta}
      \EndWhile
      \label{algo:planner:search-iteration-end}
    \end{algorithmic}
  \end{algorithm}
}

\subsection{Details}
\Cref{algo:planner} presents the pseudocode of SSSP.
Some artifacts are explained as follows.

\paragraph{Initial Roadmap Construction~(\cref{algo:planner:init-roadmap})} is done by conventional single-agent SBMP such as RRT-Connect~\cite{kuffner2000rrt}.
The objective is to secure at least one valid path from initial to goal configurations for each robot.

\paragraph{Best-first Search~(\reflines{algo:planner:init-open}{algo:planner:end-search})} realizes the core idea by maintaining a priority queue \open that stores generated nodes and a set \explored that stores already generated search situations.
For each iteration, SSSP checks whether the popped node from \open satisfies the goal condition (\cref{algo:planner:goal-check}).
If so, it returns a solution by backtracking the node.

\paragraph{Vertex Expansion~(\reflines{algo:planner:vertex-expansion}{algo:planner:vertex-expansion-end})} is implemented by steering from the target configuration to newly sampled ones, up to a fixed number $m \in \mathbb{N}_{> 0}$.
To guarantee completeness, SSSP also uses vanilla random sampling with a small probability $\lambda$ ($0.01$ in our experiments).
Each new vertex must satisfy a constraint of distance threshold, which suppresses roadmaps being too dense;
otherwise, the search space dramatically increases, making it difficult to find a solution.
Similar techniques are seen in SBMP studies~\cite{kala2013rapidly,dobson2014sparse}.
Each new vertex $q\suf{new}$ is followed by edge updates, that is, connecting $q\suf{new}$ to all vertices that $\funcname{connect}$ returns trajectories.
Note that, in practice, many sampling trials (i.e., large $m$) may compromise computation time.

\paragraph{Search Node Expansion~(\reflines{algo:planner:search-node-expansion}{algo:planner:search-node-expansion-end})} adds successors that \emph{(i)}~have not appeared yet in the search process and \emph{(ii)}~do not collide with other robots.

\paragraph{Search Iteration~(\reflines{algo:planner:search-iteration}{algo:planner:search-iteration-end}).}
The search iterates until a solution is found while decreasing the distance thresholds for vertex expansion by multiplying $0 < \gamma < 1$ by current ones.

\paragraph{Search Node Scoring.}
The heart of the best-first search is how to score each node that determines which node is popped from \open~(\cref{algo:planner:pop}).
Given a tuple of configurations \Q, SSSP scores the node by summation over each robot $i$'s shortest path distance from $\Q[i]$ to one of the configurations in $Q_i\goal$ on the roadmap $G_i$, while weighting each edge $(q\from, q\to) \in E_i$ by $\dist(q\from, q\to)$.
The path distance is calculated by ignoring inter-robot collisions, as common in heuristics for MAPF studies~\cite{silver2005cooperative}.

\subsection{Properties}
Let $k \in \mathbb{N}_{>0}$ be the number of search iterations of \reflines{algo:planner:search-iteration}{algo:planner:search-iteration-end}.
Moreover, let $G_i^k$ be the roadmap $G_i$ at the beginning of $k$-th iteration (i.e., $G_i$ at \cref{algo:planner:init-open}) and $G^k \defeq (G_i^k)^{i \in A}$.
A solution is called \emph{sequential} when at most one robot transits configurations simultaneously at any time.
\begin{lemma}
  SSSP finds a solution in the $k$-th search iteration if $G^{k}$ contains a sequential solution.
  \label{lemma:search}
\end{lemma}
\begin{proof}
  For each $k$-th iteration, $G_i$ is not infinitely increasing due to the distance threshold $\theta_i$.
  Thus, the search space is finite: $O\left(|V_1| \cdot \ldots \cdot |V_n| \cdot |A|\right)$.
  Thanks to brute-force search in finite space, SSSP finds a solution if $G^k$ contains one.
\end{proof}

\begin{theorem}
  For the geometric MRMP problem (\cref{def:mrmp}), the probability that SSSP finds a solution approaches one as $k$ approaches $\infty$, provided the instance is solvable.
  \label{thrm:completeness}
\end{theorem}
\begin{proof}
We limit the discussion to geometric MRMP wherein each robot can move in arbitrary directions in its configuration space.
The proof is based on analysis in~\cite{vsvestka1998coordinated}, which claims if an instance is solvable,
\emph{(i)}~there is a sequential solution and
\emph{(ii)}~sufficiently dense robot-wise roadmaps constructed uniformly at random sampling (e.g., PRM~\cite{kavraki1996probabilistic}) contain a sequential solution.
According to \cref{lemma:search}, SSSP finds a solution once roadmaps holding claim-(ii) are obtained.
SSSP eventually constructs such roadmaps for the following reasons.

For each search iteration, each robot $i$ tries to develop its roadmap with at least $m \geq 1$ new samples.
With the probability $1-\lambda > 0$, some of them are outcomes of uniformly at random sampling.
Each iteration terminates in finite time (see \cref{lemma:search}), therefore, each robot does not stop attempts of uniformly at random sampling until finding solutions.
Moreover, each iteration decreases the distance threshold, enabling robots to construct denser roadmaps.
\end{proof}

In short, \cref{thrm:completeness} states that SSSP eventually finds a solution.
This corresponds to \emph{probabilistic completeness} in SBMP~\cite{elbanhawi2014sampling}, defined by the probability of finding solutions bounded by the number of sampling.

{
  \begin{figure}[t!]
    \centering
    \includegraphics[width=1.0\linewidth]{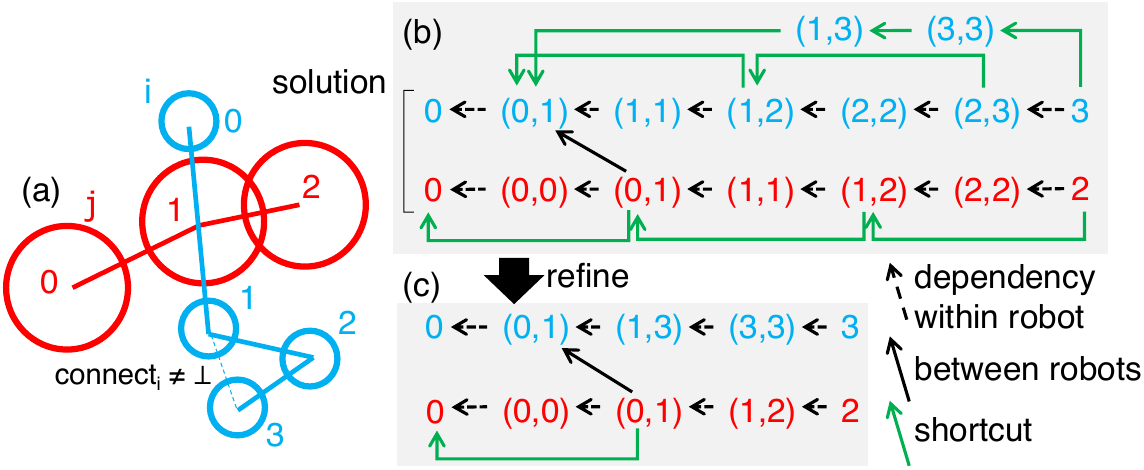}
    \caption{
      Postprocessing to refine a solution.
    }
    \label{fig:refine}
  \end{figure}
}

\subsection{Postprocessing}
\label{sec:postprocessing}
SSSP returns only sequential solutions and compromises solution quality such as the maximum traveling time.
Thus, we briefly discuss how to realize parallel execution that enables two or more robots to move simultaneously.
Specifically, we consider postprocessing to refine solutions.
Smoothing solution trajectories by postprocessing is common in single-robot SBMP~\cite{geraerts2007creating}.
However, MRMP additionally takes care of inter-robot collisions.

We describe our method with \cref{fig:refine}a.
Suppose that SSSP outputs a sequential solution of $\Pi_i = (0, 1, 1, 2, 2, 3)$ (blue) and $\Pi_j = (0, 0, 1, 1, 2, 2)$ (red).
The method repeats the next two steps until a given solution metric has not improved.
\begin{enumerate}
\item
Construct a \emph{temporal plan graph (TPG)}~\cite{honig2016multi} of the solution.
TPG is a directed acyclic graph that records temporal dependencies of each robot's motions.
We also attach possible ``shortcut'' motions to TPG.
\Cref{fig:refine}b shows an example.
There is an arc between the motions $(0,1)$ of robots $\{i,j\}$;
these motions must happen in order due to collision avoidance.
Several shortcut arcs exist, for example, $i$ can skip using vertex-$2$ by directly going from vertex-$1$ to vertex-$3$.
\item
Remove redundant motions in TPG while keeping the dependencies between robots.
\Cref{fig:refine}c shows an example.
For $j$, the motions $(1,1)$ and $(2,2)$ are removed but $(0,0)$ survives to keep the dependency with $i$.
\end{enumerate}
In the example, we finally obtain a refined solution $\Pi_i = (0, 1, 3, 3)$ and $\Pi_j = (0, 0, 1, 2)$.

The above refinement is applicable to any MRMP solutions not limited to those from SSSP.
Indeed, the experiments applied the refinement for solutions obtained by all methods.

{
  \setlength{\tabcolsep}{0pt}
  \newcommand{\figcol}[2]{
    \begin{minipage}{0.245\linewidth}
      \centering
      \begin{tabular}{cc}
        \begin{minipage}{0.54\linewidth}
          \begin{flushright}
            {\scenario{#1}}~~\\
            {\scriptsize DOF: $#2|A|$}~~
          \end{flushright}
        \end{minipage}
        &
        \begin{minipage}{0.38\linewidth}
          \IfFileExists{fig/raw/demo/#1.pdf}
          {\includegraphics[width=1.0\linewidth,left,trim={0 0.42cm 0 0.42cm},clip]
          {fig/raw/demo/#1.pdf}}{}
        \end{minipage}
      \end{tabular}\\
      \IfFileExists{fig/raw/result_#1.pdf}
      {\includegraphics[width=1.0\linewidth,trim={0 0 0 0.12cm},clip]{fig/raw/result_#1.pdf}}{}
    \end{minipage}
  }
  \begin{figure*}[th!]
    \centering
    \begin{tabular}{cccc}
      \figcol{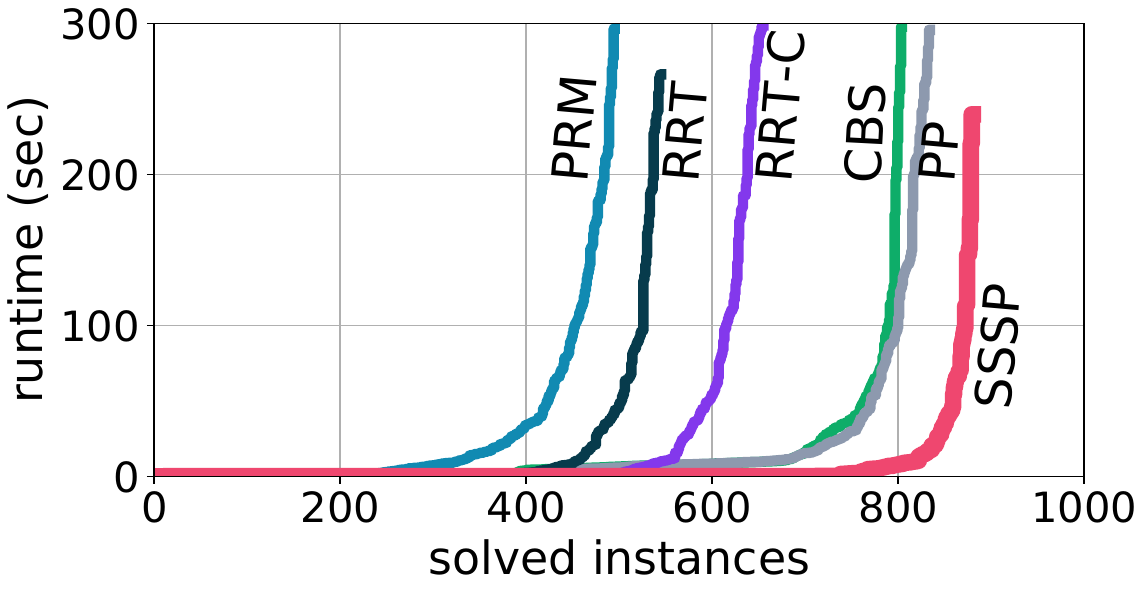}{2}
      & \figcol{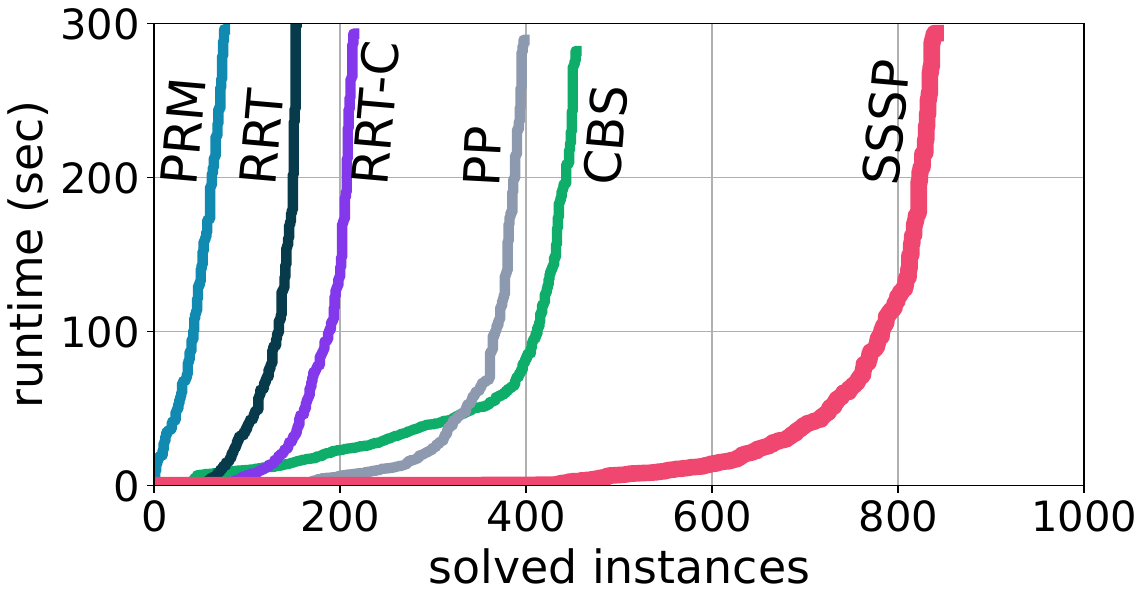}{3}
      & \figcol{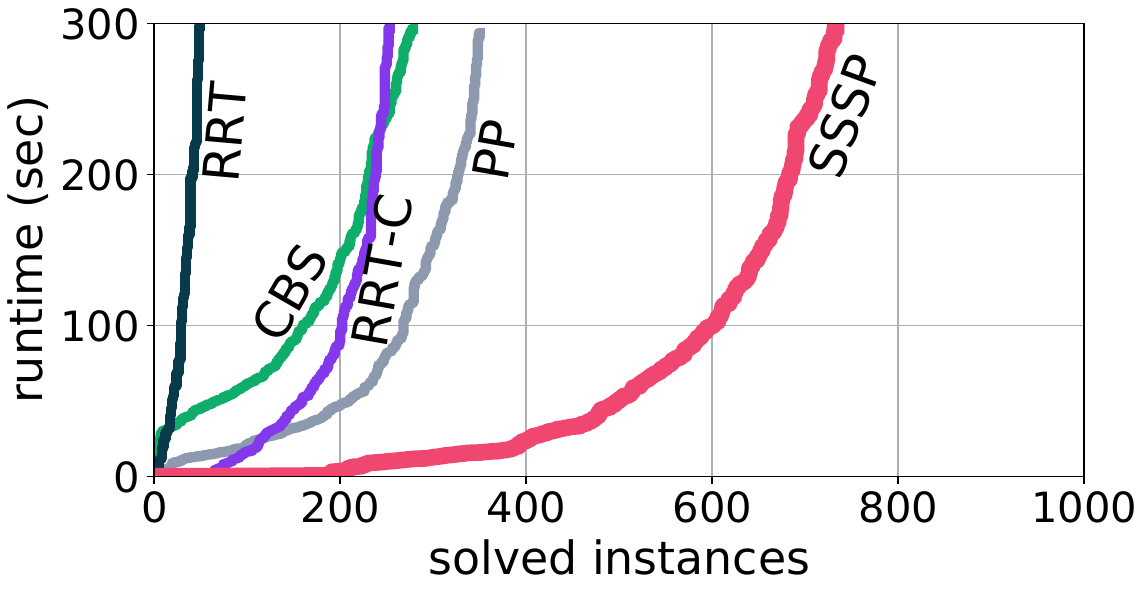}{3}
      & \figcol{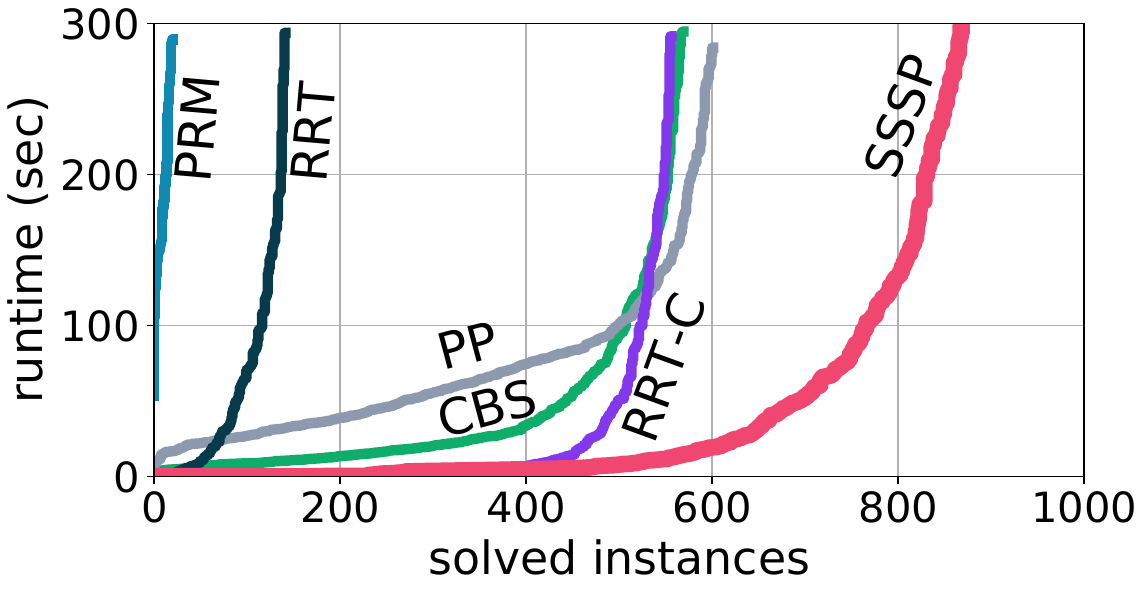}{6}
      \medskip\\
      \figcol{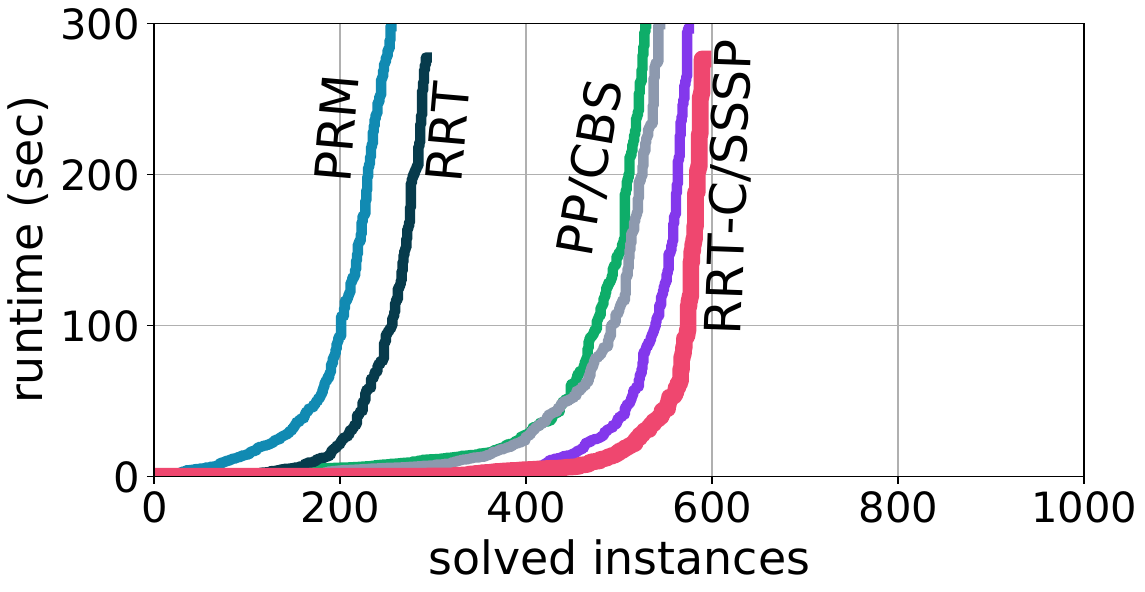}{2}
      & \figcol{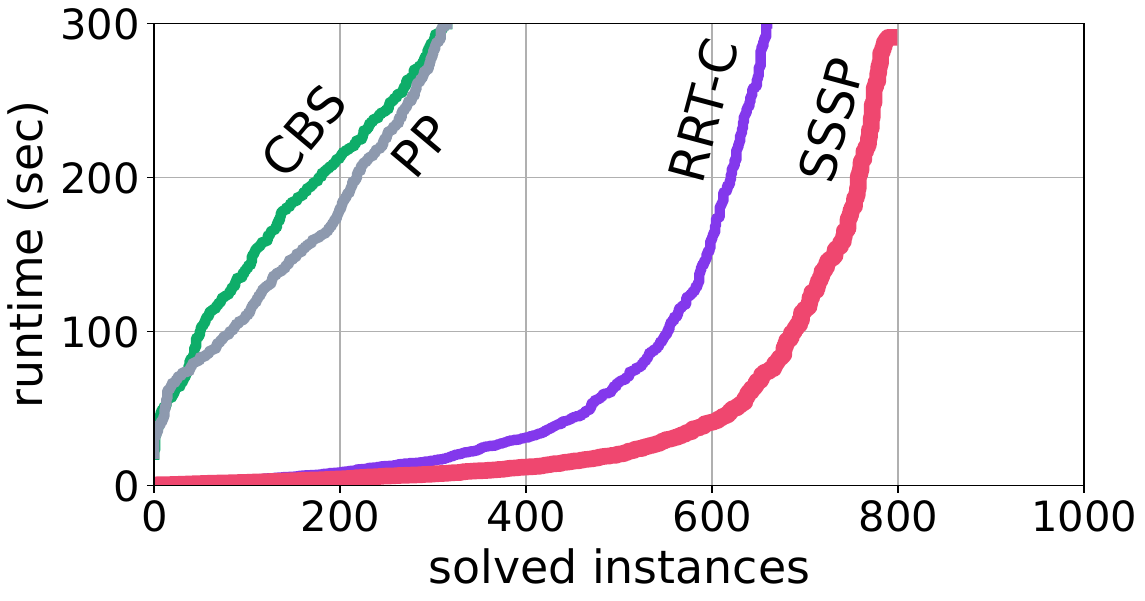}{6}
      & \figcol{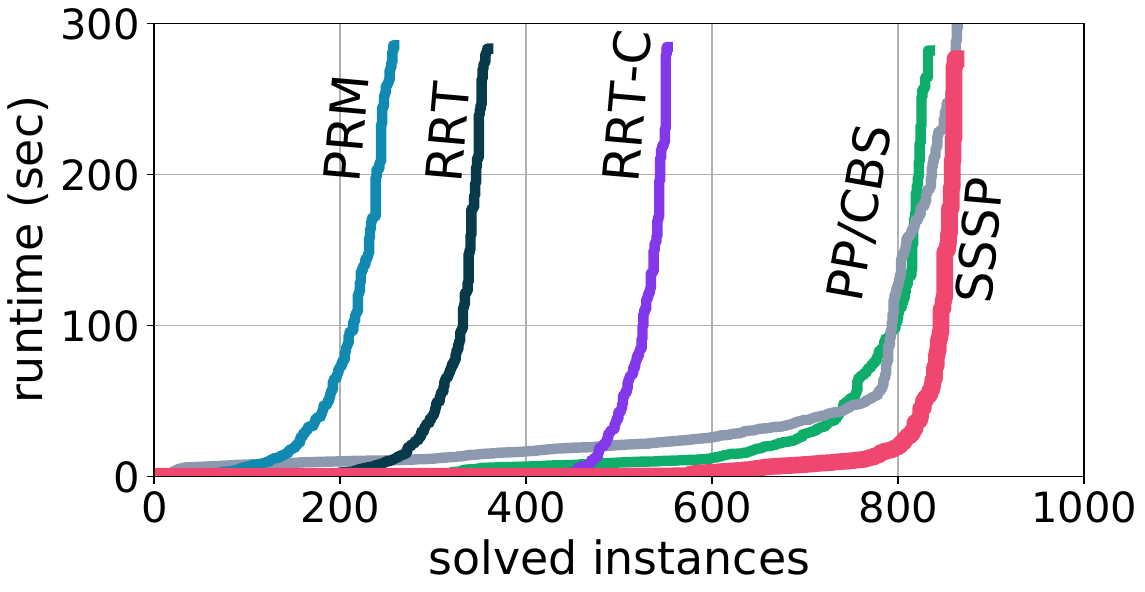}{3}
      & \figcol{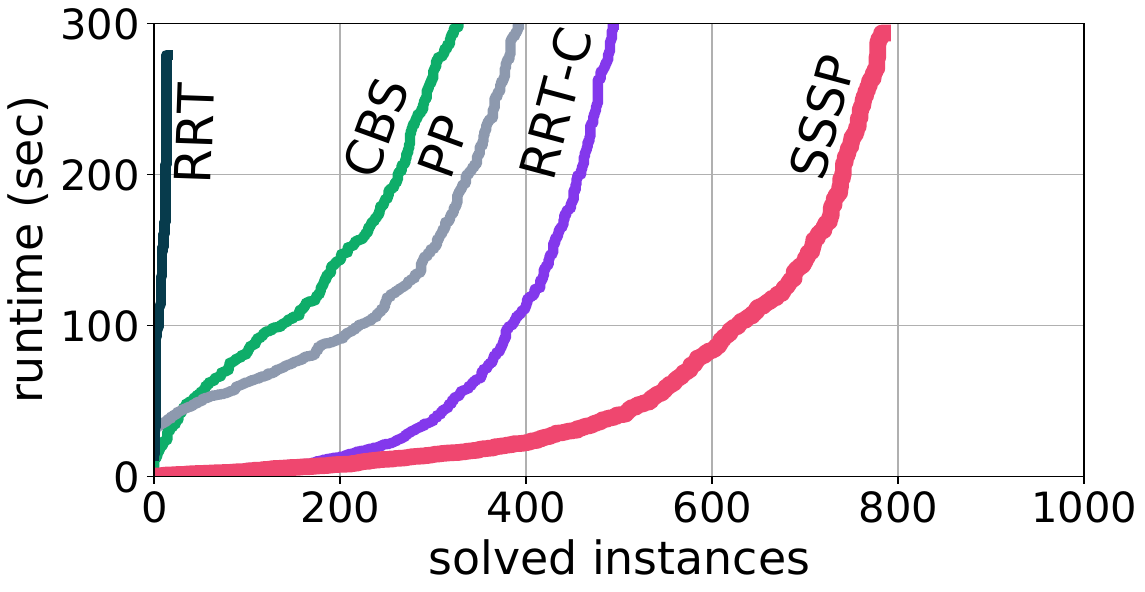}{6}
    \end{tabular}
    \caption{
      Summary of results.
      Each scenario is visualized with robots' bodies (colored circles, spheres, or lines), obstacles (black-filled circles or spheres), and a solution example (thin lines).
      Degrees of freedom (DOF) are denoted below scenario names. 
      In \scenario{arm\{22,33\}}, each robot has a fixed root, represented as colored boxes.
      These two and \scenario{snake2d} prohibit self-colliding.
      In \scenario{dubins2d}, robots must follow Dubins paths~\protect\cite{dubins1957curves}.
      The runtime scores of PP, CBS, and SSSP include the initial roadmap construction.
    }
    \label{fig:result}
  \end{figure*}
}

\section{Evaluation}
\label{sec:eval}

This section extensively evaluates SSSP on a variety of MRMP problems and demonstrates that it can solve various MRMP rapidly compared to other standard approaches.
We further assess solution quality, scalability about the number of robots $|A|$, and which components are essential for SSSP, followed by a ground-robot demo in a dense situation.

\subsection{Experimental Setups}
\label{sec:setup}
\paragraph{Benchmarks.}
As illustrated in \cref{fig:result}, we prepared various scenarios with diverse degrees of freedom and kinematic constraints, in closed workspaces $\W \in [0, 1]^{\{2, 3\}}$.
To focus on characteristics specific to MRMP, we modeled these scenarios with simple geometric patterns (e.g., spheres or lines) and reduced the effort of the \connect and \collide functions;
these functions were performed with simple geometry calculations.
For each scenario, we prepared 100 instances by randomly generating initial/goal configurations and obstacle layouts.
For each instance, the number of robots $|A|$ was chosen from the interval $\{2, 3, \ldots, 10\}$.
Robots' body parameters (e.g., radius and arm length) were also generated randomly and differed between robots.
These parameters were adjusted so that robots are sufficiently congested, otherwise, the instances become easy to solve.
Note that unsolvable instances may be included, though we excluded obviously unsolvable instances such as initial configurations with inter-robot collisions.
In summary, each instance consists of a team of heterogeneous robots and each robot has a different configuration space;
a shared roadmap is unavailable.

\paragraph{Baselines.}
Our goal is to develop planning algorithms that can be applied to various domains without the use of external knowledge other than five black-box functions.
To this end, we carefully selected the following well-known baseline methods that are applicable to MRMP defined in \cref{sec:pre}.
\footnote{
dRRT$^{(\ast)}$~\cite{solovey2016finding,shome2020drrt} was not included due to requiring an additional oracle.
}
\begin{itemize}
\item \textbf{Probabilistic roadmap (PRM)}~\cite{kavraki1996probabilistic} is a celebrated SBMP.
  For MRMP, PRM samples a composite configuration of $|A|$ robots directly from $O(|A|)$ dimensional spaces and constructs a single roadmap, and then derives a solution by pathfinding on it.
\item \textbf{Rapidly-exploring Random Tree (RRT)}~\cite{lavalle1998rapidly} is another popular SBMP, focusing on single-query situations.
  Similar to PRM, RRT for MRMP samples a composite state and constructs a tree roadmap rooted in a composite one of initial configurations for all robots.
\item \textbf{RRT-Connect (RRT-C)}~\cite{kuffner2000rrt} is a popular extension of RRT, which accelerates finding a solution by bi-directional search from both initial and goal configurations.
\item \textbf{Prioritized Planning (PP)}~\cite{erdmann1987multiple,silver2005cooperative,van2005prioritized} is a standard approach to MAPF such that robots sequentially plan paths while avoiding collisions with already planned paths.
  We applied PP to roadmaps constructed by robot-wise PRMs, as taken in~\cite{le2018cooperative}.
  PP was repeated with random priorities until the problem is solved.
\item
  \textbf{Conflict-Based Search (CBS)}~\cite{sharon2015conflict} is another popular MAPF algorithm.
  CBS is applicable to MRMP when roadmaps are given~\cite{solis2021representation}.
  We run CBS on robot-wise roadmaps constructed by PRM.
  Moreover, we manipulated the heuristic of CBS to avoid collisions as much as possible during the search.
  Doing so loses the optimality of CBS but speeds up finding solutions significantly~\cite{barer2014suboptimal}.
\end{itemize}
All hyperparameters of each method including SSSP were adjusted prior to the experiments (see the appendix).
SSSP used RRT-Connect~\cite{kuffner2000rrt} to obtain initial robot-wise roadmaps (\cref{algo:planner:init-roadmap}).
Note that this is irrelevant to RRT-C in the baselines.
PP/CBS were tested with PRM rather than RRT-Connect because otherwise constructed roadmaps do not include detours, which is essential for solving MAPF in the second phase of two-phase planning.
Since all methods rely on non-determinism, we tested each method with 10 different random seeds for each instance (1,000 trials in total).

\paragraph{Metrics.}
The objective is to find solutions as quickly as possible.
Therefore, we rate \textbf{how many instances are solved within given time limits} (maximum: \SI{5}{\minute}).

\paragraph{Evaluation Environment.}
The simulator and all methods were coded in Julia.
The experiments were run on a desktop PC with Intel Core i9-7960X \SI{2.8}{\giga\hertz} CPU and \SI{64}{\giga\byte} RAM.
A maximum of 32 different instances were run in parallel using multi-threading.
All methods used exactly the same implementations of \connect, \collide, \sample, and \dist.

\subsection{Results of Various MRMP Problems}
\label{sec:result}
\Cref{fig:result} summarizes the results.
In short, SSSP outperforms the other baselines in all tested scenarios, i.e., solving more instances much faster.
We acknowledge that runtime performance heavily relies on implementations;
however, these results indicate that SSSP is very promising.
The results of SSSP in \scenario{arm22} and \scenario{dubins2d} are relatively non-remarkable but we guess this is due to many unsolvable instances, which could be easily generated in these scenarios.
We later discuss why SSSP is quick.

\paragraph{Solution Quality.}
As reference records of solution quality, \cref{table:result-solution-quality} shows the expected \textbf{total traveling time} of all robots (aka. sum-of-costs), after applying the postprocessing introduced in \cref{sec:postprocessing} to all methods.
Compared to the other baselines, the total traveling time of SSSP is not the best but comparable.
We will discuss optimality at the end of the paper.

{
  \renewcommand{\arraystretch}{1.0}
  \setlength{\tabcolsep}{1pt}
  \newcommand{\w}[1]{\textbf{#1}}
  \newcommand{\ci}[1]{\tiny{(#1)}}
  \newcommand{\myblock}[1]{\renewcommand{\arraystretch}{0.5}\begin{tabular}{c}#1\end{tabular}}
  \newcommand{\rowhead}[2]{
    \renewcommand{\arraystretch}{0.5}
    \begin{tabular}{r}
      \scenario{#1}\\{\tiny #2}
    \end{tabular}
  }
  \begin{table}[t!]
    \centering
    \footnotesize
    \begin{tabular}{rccccccc}
      \toprule
      & SSSP
      & PRM
      & RRT
      & RRT-C
      & PP
      & CBS
      \\\midrule
      \rowhead{point2d}{54/100}
      & \myblock{1.93\\\ci{1.75, 2.10}}
      & \myblock{3.67\\\ci{2.95, 4.25}}
      & \myblock{3.42\\\ci{3.02, 3.81}}
      & \myblock{3.04\\\ci{2.62, 3.43}}
      & \myblock{\w{1.45}\\\ci{1.37, 1.52}}
      & \myblock{\w{1.43}\\\ci{1.36, 1.49}}
      \smallskip\\
      \rowhead{point3d}{26/100}
      & \myblock{2.88\\\ci{2.56, 3.16}}
      & \myblock{\NA}
      & \myblock{\NA}
      & \myblock{4.58\\\ci{4.14, 4.99}}
      & \myblock{\w{1.91}\\\ci{1.73, 2.06}}
      & \myblock{\w{1.72}\\\ci{1.59, 1.84}}
      \smallskip\\
      \rowhead{line2d}{34/100}
      & \myblock{2.68\\\ci{2.42, 2.92}}
      & \myblock{\NA}
      & \myblock{\NA}
      & \myblock{5.89\\\ci{5.14, 6.62}}
      & \myblock{1.80\\\ci{1.69, 1.90}}
      & \myblock{\w{1.62}\\\ci{1.56, 1.69}}
      \smallskip\\
      \rowhead{capsule3d}{61/100}
      & \myblock{2.34\\\ci{2.12, 2.53}}
      & \myblock{\NA}
      & \myblock{\NA}
      & \myblock{2.89\\\ci{2.55, 3.23}}
      & \myblock{\w{1.84}\\\ci{1.79, 1.90}}
      & \myblock{2.33\\\ci{2.24, 2.43}}
      \smallskip\\
      \rowhead{arm22}{30/100}
      & \myblock{\w{1.57}\\\ci{1.35, 1.75}}
      & \myblock{2.46\\\ci{2.13, 2.78}}
      & \myblock{2.23\\\ci{1.96, 2.48}}
      & \myblock{\w{1.73}\\\ci{1.43, 2.00}}
      & \myblock{\w{1.60}\\\ci{1.43, 1.73}}
      & \myblock{\w{1.51}\\\ci{1.35, 1.64}}
      \smallskip\\
      \rowhead{arm33}{94/100}
      & \myblock{\w{2.31}\\\ci{2.22, 2.40}}
      & \myblock{\NA}
      & \myblock{\NA}
      & \myblock{2.97\\\ci{2.72, 3.21}}
      & \myblock{2.74\\\ci{2.64, 2.83}}
      & \myblock{2.75\\\ci{2.68, 2.82}}
      \smallskip\\
      \rowhead{dubins2d}{30/100}
      & \myblock{1.52\\\ci{1.42, 1.62}}
      & \myblock{3.84\\\ci{3.31, 4.32}}
      & \myblock{3.06\\\ci{2.78, 3.33}}
      & \myblock{2.07\\\ci{1.80, 2.32}}
      & \myblock{1.28\\\ci{1.23, 1.32}}
      & \myblock{1.39\\\ci{1.34, 1.45}}
      \smallskip\\
      \rowhead{snake2d}{55/100}
      & \myblock{\w{3.17}\\\ci{2.91, 3.42}}
      & \myblock{\NA}
      & \myblock{\NA}
      & \myblock{4.62\\\ci{4.20, 5.02}}
      & \myblock{\w{3.28}\\\ci{3.08, 3.48}}
      & \myblock{\w{3.25}\\\ci{3.04, 3.44}}
      \\\bottomrule
    \end{tabular}
    \caption{
      Total traveling time.
      Scores are averaged over instances successfully solved by all methods, and normalized by $\sum_{i\in A}\dist(q_i\init, q \in \Q_i\goal)$.
      The numbers below on the scenario names are those numbers of instances.
      To obtain meaningful values, some cases are excluded from the calculation due to the low success rates ($\leq$20\%; denoted as \NA).
      We also show 95\% confidence intervals of means.
      Bold characters are based on overlaps of the intervals.
     }
    \label{table:result-solution-quality}
  \end{table}
}

\subsection{Scalability Test}
\label{sec:scalability}
We next assess the scalability of SSSP about the number of robots $|A|$, varied by 10 increments.
For each $|A|$, 100 \scenario{point2d} instances were prepared with smaller robots' radius (see \cref{fig:result-scalability}).
PP with tuned parameters (see the appendix) was also tested as a baseline, which relatively scored high among the other baselines in the scenario with many robots.
\Cref{fig:result-scalability} shows that, with larger $|A|$, SSSP takes longer but still acceptable time for planning, compared to the baseline.

\subsection{Which Elements are Essential?}
\label{sec:ablation}
We next address another question that asks which technical components are essential to SSSP.
Specifically, we evaluated degraded versions that omit the following components:
\emph{(i)}~initial roadmap constructions (\cref{algo:planner:init-roadmap}),
\emph{(ii)}~search node scoring (replaced by random values),
\emph{(iii)}~vertex expansion (Lines~\ref{algo:planner:vertex-expansion}--\ref{algo:planner:vertex-expansion-end})
\emph{(iv)}~distance thresholds check (\cref{algo:planner:distance-check}),
\emph{(v)}~steering (by setting $\lambda=1$), and
\emph{(vi)}~integrated sampling and search.
The last one rated SSSP without vertex expansion ($m=0$) on robot-wise PRMs.

\Cref{table:result-ablation} reveals that all these components contribute to the performance of SSSP.
Among them, involving the appropriate node scoring is particularly critical to achieving high success rates within a limited time, as well as vertex expansion.
The initial roadmap construction is effective when single-robot motion planning itself is difficult (\scenario{snake2d}).
The steering effect was non-dramatic because the workspace is small relative to robots;
we discuss this in the appendix, together with scores of total traveling time.

{
  \setlength{\tabcolsep}{0pt}
  \begin{figure}[t!]
    \centering
    \begin{tabular}{cc}
      \begin{minipage}{0.33\linewidth}
        \centering
        \includegraphics[width=1.0\linewidth,left]{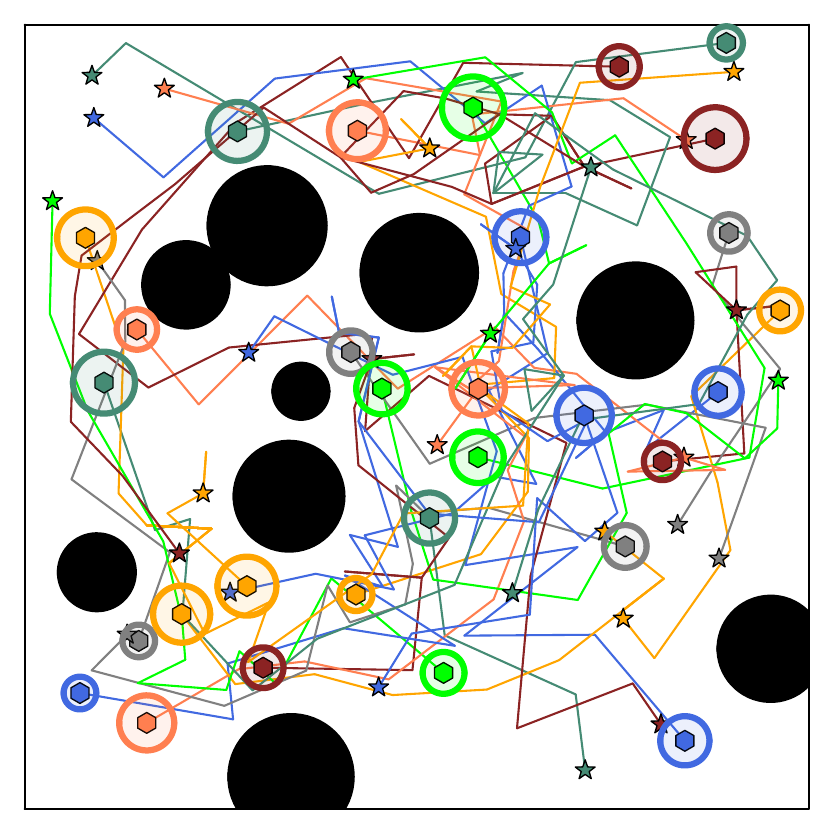}
      \end{minipage}
      &
        \begin{minipage}{0.66\linewidth}
          \centering
          \includegraphics[width=1.0\linewidth,left]{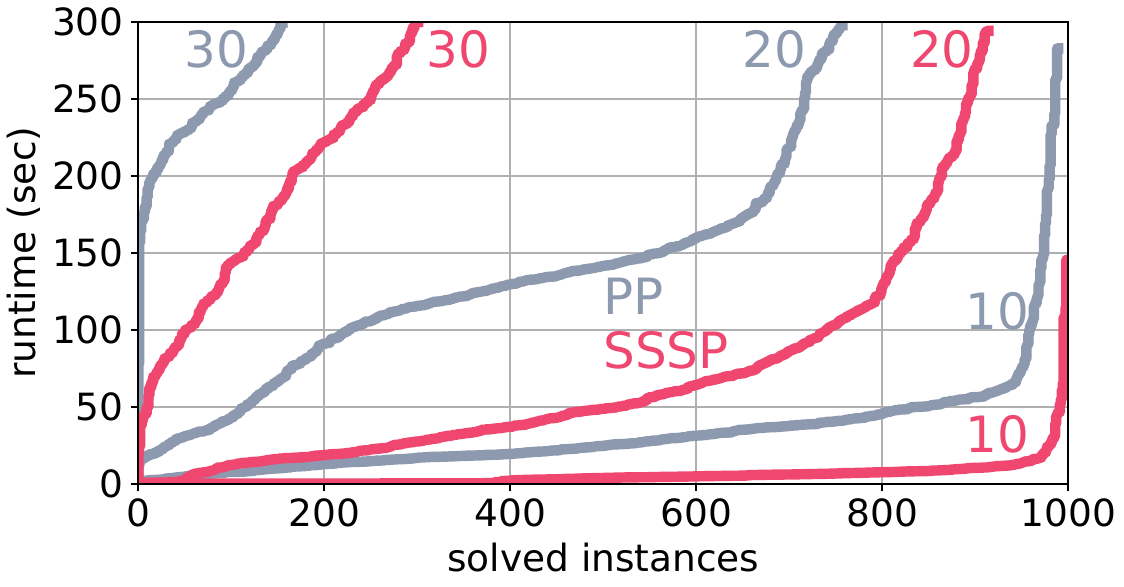}
        \end{minipage}

    \end{tabular}
    \caption{
      Scalability test in \scenario{point2d}.
      \emph{left}:~An example instance ($|A|=30$) and its solution from SSSP.
      \emph{right}:~The number of solved instances with annotation of $|A|$.
    }
    \label{fig:result-scalability}
  \end{figure}
}

{
  \setlength{\tabcolsep}{0.7pt}
  \newcommand{\w}[1]{\textbf{#1}}
  \newcommand{\head}[1]{
    {\renewcommand{\arraystretch}{0.8}\begin{tabular}{c}#1\end{tabular}}
  }
  \begin{table}[t!]
    \centering
    \footnotesize
    \begin{tabular}{rccccccc}
      \toprule
      & SSSP
      & \head{random\\score}
      & \head{no init\\roadmap}
      & \head{no vertex\\expansion}
      & \head{no \dist\\check}
      & \head{$\lambda$$=$$1$}
      & \head{on\\PRM}
      \\\midrule
      \scenario{point2d}
      & \w{880} & 512 & 793 & 498 & 737 & 875 & 807
      \smallskip\\
      \scenario{arm22}
      & \w{586} & 101 & 565 & 388 & 523 & 584 & 398
      \smallskip\\
      \scenario{snake2d}
      & \w{710} & 0 & 379 & 674 & 519 & 677 & 39
      \\\bottomrule
    \end{tabular}
    \caption{
      Ablation study.
      The numbers of instances solved within \SI{5}{\minute} are shown.
    }
    \label{table:result-ablation}
  \end{table}
}

{
  \setlength{\tabcolsep}{0.5pt}
  \newcommand{\entry}[1]{
    \begin{minipage}{0.32\linewidth}
      \centering
      \includegraphics[width=1\linewidth]{fig/raw/robot/#1_reduced}
    \end{minipage}
  }
  \begin{figure}[t!]
  \centering
    \begin{tabular}{ccc}
    \entry{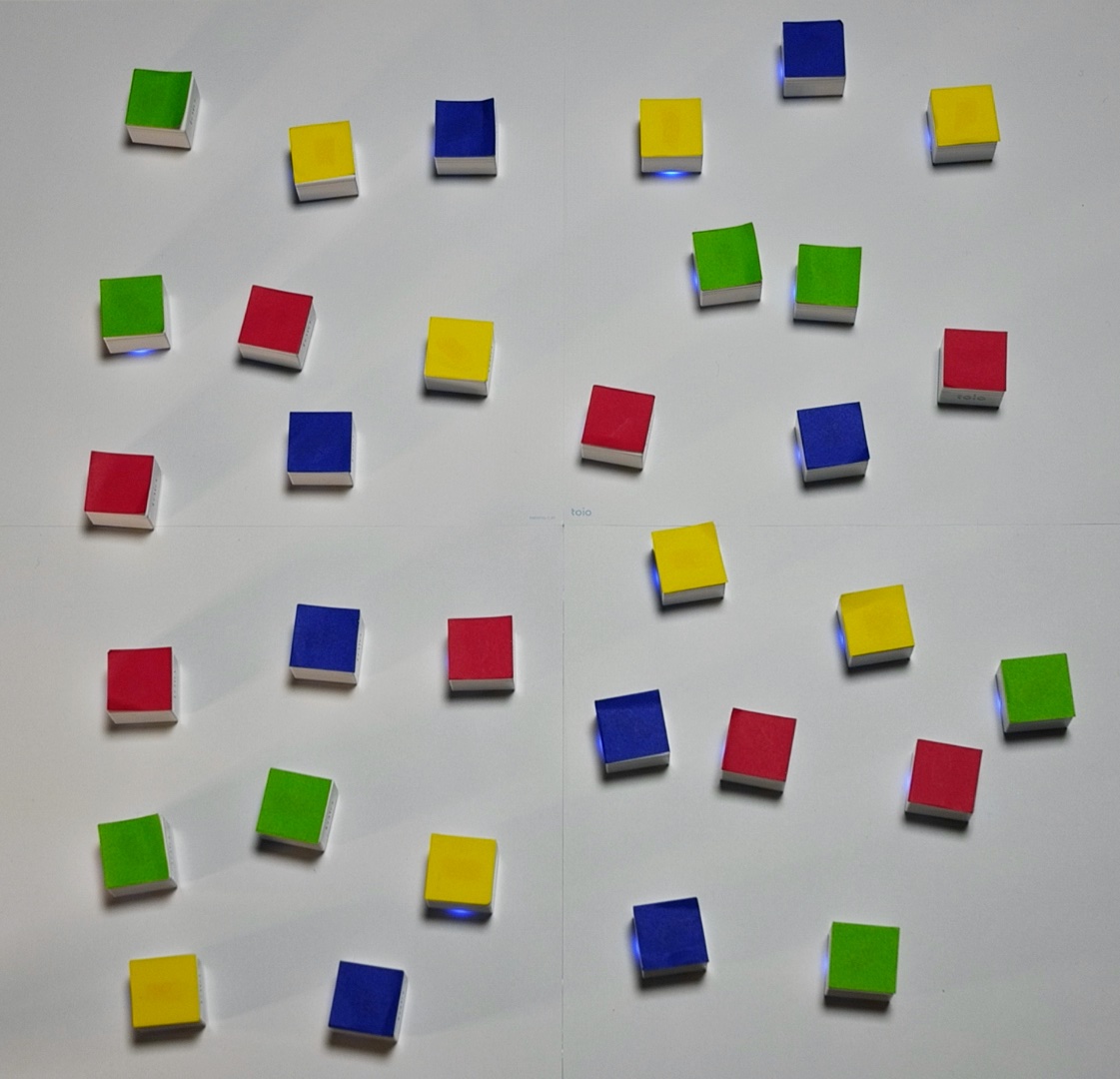}
    &
    \entry{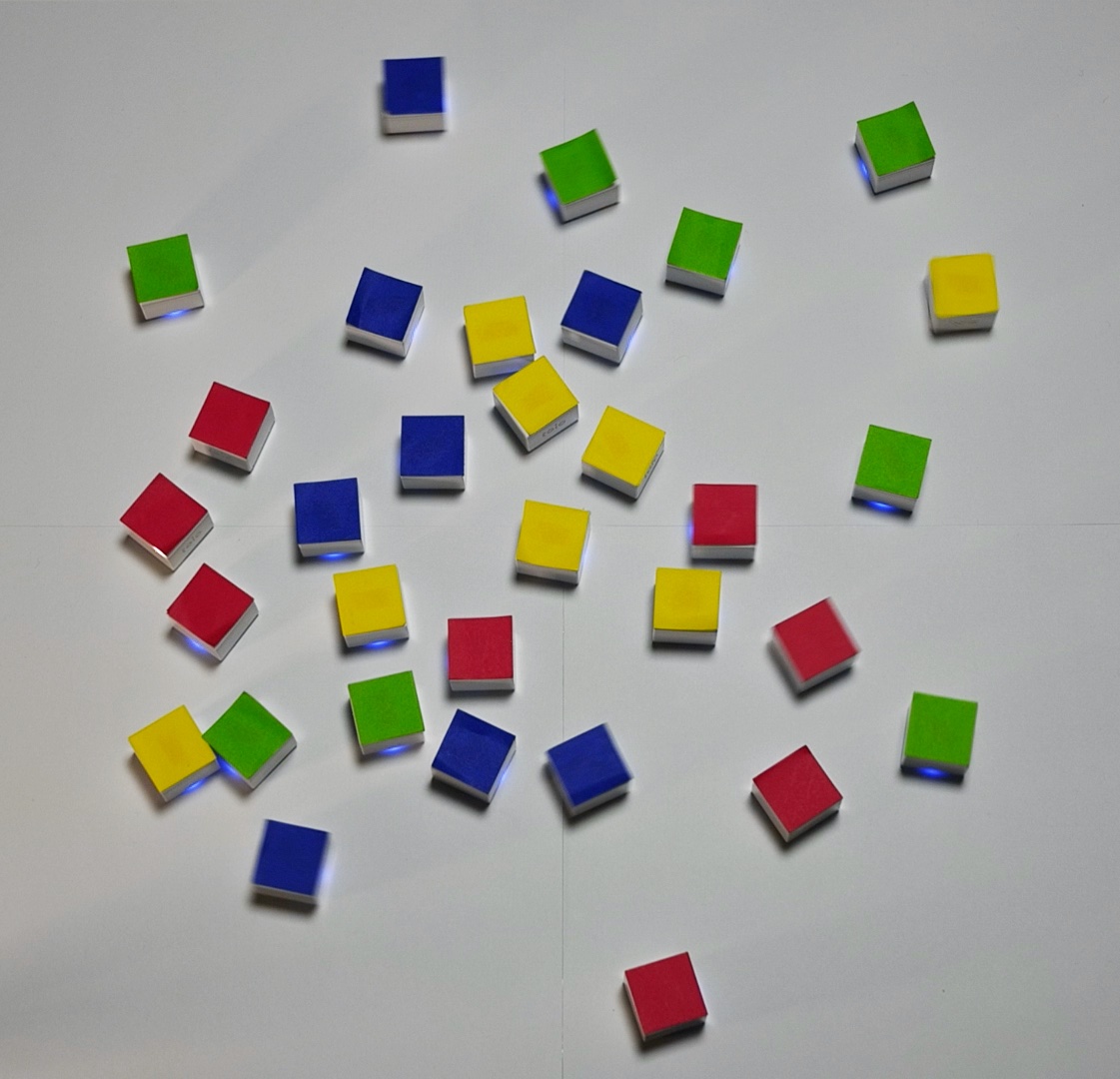}
    &
    \entry{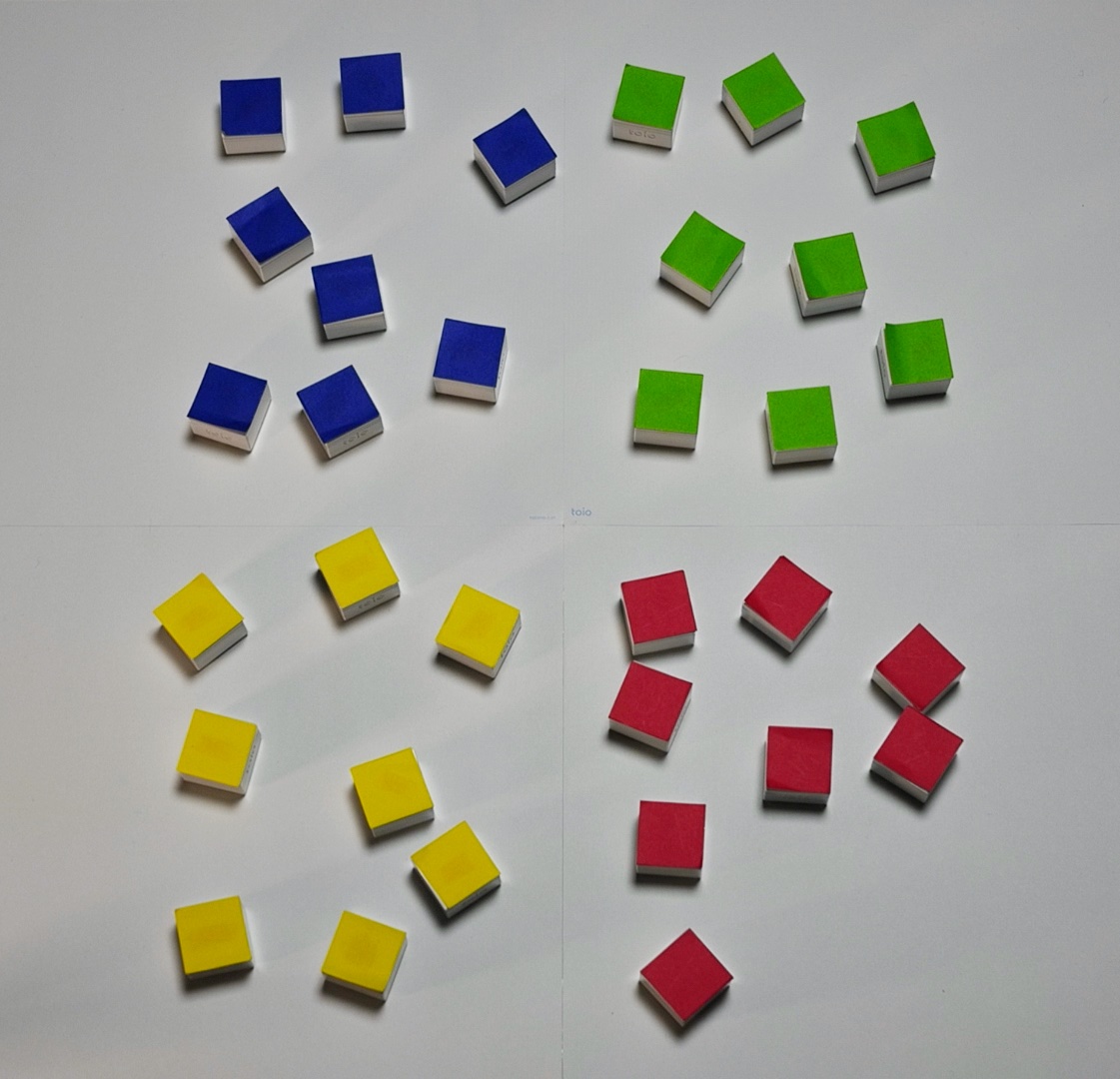}
    \end{tabular}
    \caption{
      Robot demo.
      From left to right, the pictures show the initial, intermediate, and final configurations, which eventually constitute the characters `SSSP.'
    }
    \label{fig:demo}
  \end{figure}
}

\subsection{Robot Demo}
\label{sec:robot-demo}
We applied SSSP to 32 robots (\url{https://toio.io/}) modeled as \scenario{point2d} in a dense situation (\cref{fig:demo}).
The robots evolve on a specific playmat and are controllable by instructions of absolute coordinates.
Even though an experimenter randomly placed the robots as the initial states (see the movie), the planning was done in about \SI{30}{\second}, and then all robots eventually reached their goal.
Importantly, this demo was based on only five utility functions of \connect, \sample, \dist, \steer, and \collide, without any prepared environmental representations.

\section{Conclusion and Discussion}
\label{sec:discussion}
This paper introduced the SSSP algorithm that rapidly solves MRMP.
The main idea behind the algorithm was to unite techniques developed for SBMP and search techniques for MAPF.
The former had been studied mainly in the robotics community, while the latter in the AI community.
Bringing them together brings promising results, as extensively demonstrated in our experiments.
In the remainder, we provide further discussions and future directions.

\paragraph{Why is SSSP Quick?}
We provide two qualitative explanations: one from the MAPF side, and another from the SBMP side.
The quickness of SSSP relies on both factors.

\emph{Branching Factor:}
During the search, SSSP decomposes successors into search nodes corresponding to at most one robot taking a motion.
Compared to search styles allowing all robots to move simultaneously, the decomposition significantly reduces \emph{branching factor}, i.e., the number of successors at each node.
In general, the average branching factor $b$ largely determines the search effort~\cite{edelkamp2011heuristic}.
Assume that each robot has $k$ possible motions from each configuration on average.
Coupled with perfect heuristics and a perfect tie-breaking strategy, allowing all robots to move simultaneously results in $b = k^{|A|}$ and generates $\left(k^{|A|}\right)\cdot l$ nodes, where $l$ is the depth of the search.
In contrast, SSSP results in $b = k$ and enables searching the equivalent node with only $(k|A|)\cdot l$ nodes generation.
This is a key trick of \astar with operator decomposition~\cite{standley2010finding} for MAPF;
we based this idea to develop SSSP.

\emph{Imbalanced Roadmaps:}
PRM-based methods (i.e., PRM, PP, and CBS in our experiments) have no choice other than to construct roadmaps uniformly spread in each configuration space, making search spaces huge.
Such drawbacks might be relieved with biased sampling but representing good bias for MRMP is not trivial;
indeed, existing studies use machine learning~\cite{arias2021avoidance,okumura2022ctrm}.
In contrast, owing to carefully-designed components, SSSP naturally constructs sparse roadmaps in important regions for each robot as seen in \cref{fig:roadmap-example}.
Consequently, the search space for SSSP is kept small, which also contributes to quick MRMP.

\paragraph{Kinodynamic MRMP.}
We untreated dynamics constraints, therefore, \emph{kinodynamic MRMP} is an interesting direction.
Similarly to RRT~\cite{lavalle2001randomized} or EST~\cite{hsu1997path} that are applicable to kinodynamic planning, we consider that SSSP is also applicable to such planning.
The adaptation is by considering planning with a \emph{state} $x$ instead of a configuration $q \in \C_i$ for robot $i$, which comprises a configuration $q$ and its derivative $\dot{q}$ (i.e., $x \defeq (q, \dot{q})$).
Some parts require care;
in usual kinodynamic MRMP, $\connect_i(x, x) = \bot$ as we see in cars that cannot stop instantly.
This means that sequential solutions are not allowed.
In this case, it is necessary to regard $|A|$ successive search nodes in SSSP as ``one block'' to enable concurrent motions of multiple robots.

\paragraph{Optimal MRMP.}
SSSP prioritizes solving MRMP itself, rather than solution quality.
However, it is possible to reflect quality by modifying node scoring, which is currently designed as a greedy search.
With terminologies of \astar search~\cite{hart1968formal}, SSSP only uses h-value (i.e., estimation of cost-to-go).
A promising direction is to incorporate g-value (i.e., cost-to-come), letting SSSP asymptotically optimal without degrading the performance of solvability.
We also point out that, to be optimal, rewiring the search tree is mandatory as seen in SBMP or MAPF studies~\cite{karaman2011sampling,okumura2023lacam2}.
Another possibility is to incorporate iterative refinement schemes~\cite{okumura2021iterative,li2021anytime} developed for MAPF, although there is no theoretical guarantee.

\paragraph{Further Integration of SBMP and MAPF.}
The concept behind the paper was developing robot-wise roadmaps according to the multi-agent search progress, turning out to be promising.
We consider this direction should be further investigated.
A very-recent study~\cite{kottinger2022conflict} explores this direction for CBS.
Many powerful MAPF algorithms exist not limited to \astar adaptation or CBS, such as~\cite{li2022mapf,okumura2023lacam}.
Therefore, integrating them with SBMP may fruit practical methodologies for MRMP.

\section*{Acknowledgments}
We thank Fran\c{c}ois Bonnet and the anonymous reviewers for their many insightful comments on the manuscript.
This work was partly supported by JSPS KAKENHI Grant Number 20J23011, 21K11748, 21H03423, 20K11685 and JST ACT-X Grant Number JPMJAX22A1.
Keisuke Okumura thanks the support of the Yoshida Scholarship Foundation when he was a Ph.D. student.

\bibliographystyle{named}
\bibliography{ref-macro,ref}

\begin{thebibliography}{}

\bibitem[\protect\citeauthoryear{Arias \bgroup \em et al.\egroup
  }{2021}]{arias2021avoidance}
Felipe~Felix Arias, Brian Ichter, Aleksandra Faust, and Nancy~M Amato.
\newblock Avoidance critical probabilistic roadmaps for motion planning in
  dynamic environments.
\newblock In {\em Proceedings of IEEE International Conference on Robotics and
  Automation (ICRA)}, 2021.

\bibitem[\protect\citeauthoryear{Barer \bgroup \em et al.\egroup
  }{2014}]{barer2014suboptimal}
Max Barer, Guni Sharon, Roni Stern, and Ariel Felner.
\newblock Suboptimal variants of the conflict-based search algorithm for the
  multi-agent pathfinding problem.
\newblock In {\em Proceedings of Annual Symposium on Combinatorial Search
  (SOCS)}, 2014.

\bibitem[\protect\citeauthoryear{C{\'a}p \bgroup \em et al.\egroup
  }{2013}]{cap2013multi}
Michal C{\'a}p, P.~Nov{\'a}k, J.~Vokr{\'i}nek, and M.~Pechoucek.
\newblock Multi-agent rrt: Sampling-based cooperative pathfinding.
\newblock In {\em Proceedings of International Joint Conference on Autonomous
  Agents \& Multiagent Systems (AAMAS)}, 2013.

\bibitem[\protect\citeauthoryear{Choset \bgroup \em et al.\egroup
  }{2005}]{choset2005principles}
Howie Choset, Kevin~M Lynch, Seth Hutchinson, George~A Kantor, and Wolfram
  Burgard.
\newblock {\em Principles of robot motion: theory, algorithms, and
  implementations}.
\newblock MIT press, 2005.

\bibitem[\protect\citeauthoryear{Cohen \bgroup \em et al.\egroup
  }{2019}]{cohen2019optimal}
Liron Cohen, Tansel Uras, TK~Satish Kumar, and Sven Koenig.
\newblock Optimal and bounded-suboptimal multi-agent motion planning.
\newblock In {\em Proceedings of Annual Symposium on Combinatorial Search
  (SOCS)}, 2019.

\bibitem[\protect\citeauthoryear{Dayan \bgroup \em et al.\egroup
  }{2021}]{dayan2021near}
Dror Dayan, Kiril Solovey, Marco Pavone, and Dan Halperin.
\newblock Near-optimal multi-robot motion planning with finite sampling.
\newblock In {\em Proceedings of IEEE International Conference on Robotics and
  Automation (ICRA)}, 2021.

\bibitem[\protect\citeauthoryear{Dobson and Bekris}{2014}]{dobson2014sparse}
Andrew Dobson and Kostas~E Bekris.
\newblock Sparse roadmap spanners for asymptotically near-optimal motion
  planning.
\newblock {\em International Journal of Robotics Research (IJRR)}, 2014.

\bibitem[\protect\citeauthoryear{Donald \bgroup \em et al.\egroup
  }{1993}]{donald1993kinodynamic}
Bruce Donald, Patrick Xavier, John Canny, and John Reif.
\newblock Kinodynamic motion planning.
\newblock {\em Journal of the ACM (JACM)}, 1993.

\bibitem[\protect\citeauthoryear{Dubins}{1957}]{dubins1957curves}
Lester~E Dubins.
\newblock On curves of minimal length with a constraint on average curvature,
  and with prescribed initial and terminal positions and tangents.
\newblock {\em American Journal of Mathematics}, 1957.

\bibitem[\protect\citeauthoryear{Edelkamp and
  Schrodl}{2011}]{edelkamp2011heuristic}
Stefan Edelkamp and Stefan Schrodl.
\newblock Introduction.
\newblock In {\em Heuristic search: theory and applications}, chapter~1. 2011.

\bibitem[\protect\citeauthoryear{Elbanhawi and
  Simic}{2014}]{elbanhawi2014sampling}
Mohamed Elbanhawi and Milan Simic.
\newblock Sampling-based robot motion planning: A review.
\newblock {\em Ieee access}, 2014.

\bibitem[\protect\citeauthoryear{Erdmann and
  Lozano-Perez}{1987}]{erdmann1987multiple}
Michael Erdmann and Tomas Lozano-Perez.
\newblock On multiple moving objects.
\newblock {\em Algorithmica}, 1987.

\bibitem[\protect\citeauthoryear{Feng \bgroup \em et al.\egroup
  }{2020}]{feng2020overview}
Zhi Feng, Guoqiang Hu, Yajuan Sun, and Jeffrey Soon.
\newblock An overview of collaborative robotic manipulation in multi-robot
  systems.
\newblock {\em Annual Reviews in Control}, 2020.

\bibitem[\protect\citeauthoryear{Geraerts and
  Overmars}{2007}]{geraerts2007creating}
Roland Geraerts and Mark~H Overmars.
\newblock Creating high-quality paths for motion planning.
\newblock {\em International Journal of Robotics Research (IJRR)}, 2007.

\bibitem[\protect\citeauthoryear{Han \bgroup \em et al.\egroup
  }{2018}]{han2018sear}
Shuai~D Han, Edgar~J Rodriguez, and Jingjin Yu.
\newblock Sear: A polynomial-time multi-robot path planning algorithm with
  expected constant-factor optimality guarantee.
\newblock In {\em Proceedings of IEEE/RSJ International Conference on
  Intelligent Robots and Systems (IROS)}, 2018.

\bibitem[\protect\citeauthoryear{Hart \bgroup \em et al.\egroup
  }{1968}]{hart1968formal}
Peter~E Hart, Nils~J Nilsson, and Bertram Raphael.
\newblock A formal basis for the heuristic determination of minimum cost paths.
\newblock {\em IEEE transactions on Systems Science and Cybernetics}, 1968.

\bibitem[\protect\citeauthoryear{Hearn and Demaine}{2005}]{hearn2005pspace}
Robert~A Hearn and Erik~D Demaine.
\newblock Pspace-completeness of sliding-block puzzles and other problems
  through the nondeterministic constraint logic model of computation.
\newblock {\em Theoretical Computer Science (TCS)}, 2005.

\bibitem[\protect\citeauthoryear{H{\"o}nig \bgroup \em et al.\egroup
  }{2016}]{honig2016multi}
Wolfgang H{\"o}nig, TK~Satish Kumar, Liron Cohen, Hang Ma, Hong Xu, Nora
  Ayanian, and Sven Koenig.
\newblock Multi-agent path finding with kinematic constraints.
\newblock In {\em Proceedings of International Conference on Automated Planning
  and Scheduling (ICAPS)}, 2016.

\bibitem[\protect\citeauthoryear{H{\"o}nig \bgroup \em et al.\egroup
  }{2018}]{honig2018trajectory}
Wolfgang H{\"o}nig, James~A Preiss, TK~Satish Kumar, Gaurav~S Sukhatme, and
  Nora Ayanian.
\newblock Trajectory planning for quadrotor swarms.
\newblock {\em IEEE Transactions on Robotics (T-RO)}, 2018.

\bibitem[\protect\citeauthoryear{Hopcroft \bgroup \em et al.\egroup
  }{1984}]{hopcroft1984complexity}
John~E Hopcroft, Jacob~Theodore Schwartz, and Micha Sharir.
\newblock On the complexity of motion planning for multiple independent
  objects; pspace-hardness of the warehouseman's problem.
\newblock {\em International Journal of Robotics Research (IJRR)}, 1984.

\bibitem[\protect\citeauthoryear{Hsu \bgroup \em et al.\egroup
  }{1997}]{hsu1997path}
David Hsu, J-C Latombe, and Rajeev Motwani.
\newblock Path planning in expansive configuration spaces.
\newblock In {\em Proceedings of IEEE International Conference on Robotics and
  Automation (ICRA)}, 1997.

\bibitem[\protect\citeauthoryear{Kala}{2013}]{kala2013rapidly}
Rahul Kala.
\newblock Rapidly exploring random graphs: motion planning of multiple mobile
  robots.
\newblock {\em Advanced Robotics}, 2013.

\bibitem[\protect\citeauthoryear{Karaman and
  Frazzoli}{2011}]{karaman2011sampling}
Sertac Karaman and Emilio Frazzoli.
\newblock Sampling-based algorithms for optimal motion planning.
\newblock {\em International Journal of Robotics Research (IJRR)}, 2011.

\bibitem[\protect\citeauthoryear{Kavraki \bgroup \em et al.\egroup
  }{1996}]{kavraki1996probabilistic}
Lydia~E Kavraki, Petr Svestka, J-C Latombe, and Mark~H Overmars.
\newblock Probabilistic roadmaps for path planning in high-dimensional
  configuration spaces.
\newblock {\em IEEE Transactions on Robotics and Automation}, 1996.

\bibitem[\protect\citeauthoryear{Kottinger \bgroup \em et al.\egroup
  }{2022}]{kottinger2022conflict}
Justin Kottinger, Shaull Almagor, and Morteza Lahijanian.
\newblock Conflict-based search for multi-robot motion planning with
  kinodynamic constraints.
\newblock In {\em Proceedings of IEEE/RSJ International Conference on
  Intelligent Robots and Systems (IROS)}, 2022.

\bibitem[\protect\citeauthoryear{Kuffner and LaValle}{2000}]{kuffner2000rrt}
James~J Kuffner and Steven~M LaValle.
\newblock Rrt-connect: An efficient approach to single-query path planning.
\newblock In {\em Proceedings of IEEE International Conference on Robotics and
  Automation (ICRA)}, 2000.

\bibitem[\protect\citeauthoryear{LaValle and
  Kuffner~Jr}{2001}]{lavalle2001randomized}
Steven~M LaValle and James~J Kuffner~Jr.
\newblock Randomized kinodynamic planning.
\newblock {\em International Journal of Robotics Research (IJRR)}, 2001.

\bibitem[\protect\citeauthoryear{LaValle}{1998}]{lavalle1998rapidly}
Steven~M LaValle.
\newblock Rapidly-exploring random trees: A new tool for path planning.
\newblock Technical report, Computer Science Department, Iowa State University
  (TR 98–11), 1998.

\bibitem[\protect\citeauthoryear{LaValle}{2006}]{lavalle2006planning}
Steven~M LaValle.
\newblock {\em Planning algorithms}.
\newblock Cambridge University Press, 2006.

\bibitem[\protect\citeauthoryear{Le and Plaku}{2018}]{le2018cooperative}
Duong Le and Erion Plaku.
\newblock Cooperative, dynamics-based, and abstraction-guided multi-robot
  motion planning.
\newblock {\em Journal of Artificial Intelligence Research (JAIR)}, 2018.

\bibitem[\protect\citeauthoryear{Li \bgroup \em et al.\egroup
  }{2021}]{li2021anytime}
Jiaoyang Li, Zhe Chen, Daniel Harabor, P~Stuckey, and Sven Koenig.
\newblock Anytime multi-agent path finding via large neighborhood search.
\newblock In {\em Proceedings of International Joint Conference on Artificial
  Intelligence (IJCAI)}, 2021.

\bibitem[\protect\citeauthoryear{Li \bgroup \em et al.\egroup
  }{2022}]{li2022mapf}
Jiaoyang Li, Zhe Chen, Daniel Harabor, Peter~J Stuckey, and Sven Koenig.
\newblock Mapf-lns2: Fast repairing for multi-agent path finding via large
  neighborhood search.
\newblock In {\em Proceedings of AAAI Conference on Artificial Intelligence
  (AAAI)}, 2022.

\bibitem[\protect\citeauthoryear{Okumura \bgroup \em et al.\egroup
  }{2021}]{okumura2021iterative}
Keisuke Okumura, Yasumasa Tamura, and Xavier D\'{e}fago.
\newblock Iterative refinement for real-time multi-robot path planning.
\newblock In {\em Proceedings of IEEE/RSJ International Conference on
  Intelligent Robots and Systems (IROS)}, 2021.

\bibitem[\protect\citeauthoryear{Okumura \bgroup \em et al.\egroup
  }{2022}]{okumura2022ctrm}
Keisuke Okumura, Ryo Yonetani, Mai Nishimura, and Asako Kanezaki.
\newblock Ctrms: Learning to construct cooperative timed roadmaps for
  multi-agent path planning in continuous spaces.
\newblock In {\em Proceedings of International Joint Conference on Autonomous
  Agents \& Multiagent Systems (AAMAS)}, 2022.

\bibitem[\protect\citeauthoryear{Okumura}{2023a}]{okumura2023lacam2}
Keisuke Okumura.
\newblock Improving lacam for scalable eventually optimal multi-agent
  pathfinding.
\newblock In {\em Proceedings of International Joint Conference on Artificial
  Intelligence (IJCAI)}, 2023.

\bibitem[\protect\citeauthoryear{Okumura}{2023b}]{okumura2023lacam}
Keisuke Okumura.
\newblock Lacam: Search-based algorithm for quick multi-agent pathfinding.
\newblock In {\em Proceedings of AAAI Conference on Artificial Intelligence
  (AAAI)}, 2023.

\bibitem[\protect\citeauthoryear{Pan \bgroup \em et al.\egroup
  }{2012}]{pan2012fcl}
Jia Pan, Sachin Chitta, and Dinesh Manocha.
\newblock Fcl: A general purpose library for collision and proximity queries.
\newblock In {\em Proceedings of IEEE International Conference on Robotics and
  Automation (ICRA)}, 2012.

\bibitem[\protect\citeauthoryear{Reif}{1979}]{reif1979complexity}
John~H Reif.
\newblock Complexity of the mover's problem and generalizations.
\newblock In {\em Proceedings of Annual Symposium on Foundations of Computer
  Science (FOCS)}, 1979.

\bibitem[\protect\citeauthoryear{S{\'a}nchez and
  Latombe}{2002}]{sanchez2002delaying}
Gildardo S{\'a}nchez and Jean-Claude Latombe.
\newblock On delaying collision checking in prm planning: Application to
  multi-robot coordination.
\newblock {\em International Journal of Robotics Research (IJRR)}, 2002.

\bibitem[\protect\citeauthoryear{Sharon \bgroup \em et al.\egroup
  }{2015}]{sharon2015conflict}
Guni Sharon, Roni Stern, Ariel Felner, and Nathan~R Sturtevant.
\newblock Conflict-based search for optimal multi-agent pathfinding.
\newblock {\em Artificial Intelligence (AIJ)}, 2015.

\bibitem[\protect\citeauthoryear{Shome \bgroup \em et al.\egroup
  }{2020}]{shome2020drrt}
Rahul Shome, Kiril Solovey, Andrew Dobson, Dan Halperin, and Kostas~E Bekris.
\newblock drrt\textasteriskcentered: Scalable and informed
  asymptotically-optimal multi-robot motion planning.
\newblock {\em Autonomous Robots (AURO)}, 2020.

\bibitem[\protect\citeauthoryear{Silver}{2005}]{silver2005cooperative}
David Silver.
\newblock Cooperative pathfinding.
\newblock In {\em Proceedings of AAAI Conference on Artificial Intelligence and
  Interactive Digital Entertainment (AIIDE)}, 2005.

\bibitem[\protect\citeauthoryear{Solis \bgroup \em et al.\egroup
  }{2021}]{solis2021representation}
Irving Solis, James Motes, Read Sandstr{\"o}m, and Nancy~M Amato.
\newblock Representation-optimal multi-robot motion planning using
  conflict-based search.
\newblock {\em IEEE Robotics and Automation Letters (RA-L)}, 2021.

\bibitem[\protect\citeauthoryear{Solovey \bgroup \em et al.\egroup
  }{2016}]{solovey2016finding}
Kiril Solovey, Oren Salzman, and Dan Halperin.
\newblock Finding a needle in an exponential haystack: Discrete rrt for
  exploration of implicit roadmaps in multi-robot motion planning.
\newblock {\em International Journal of Robotics Research (IJRR)}, 2016.

\bibitem[\protect\citeauthoryear{Spirakis and Yap}{1984}]{spirakis1984np}
P.~Spirakis and C.~Yap.
\newblock Strong np-hardness of moving many discs.
\newblock {\em Information Processing Letters}, 1984.

\bibitem[\protect\citeauthoryear{Standley}{2010}]{standley2010finding}
Trevor~Scott Standley.
\newblock Finding optimal solutions to cooperative pathfinding problems.
\newblock In {\em Proceedings of AAAI Conference on Artificial Intelligence
  (AAAI)}, 2010.

\bibitem[\protect\citeauthoryear{Stern \bgroup \em et al.\egroup
  }{2019}]{stern2019def}
Roni Stern, Nathan Sturtevant, Ariel Felner, Sven Koenig, Hang Ma, Thayne
  Walker, Jiaoyang Li, Dor Atzmon, Liron Cohen, TK~Kumar, et~al.
\newblock Multi-agent pathfinding: Definitions, variants, and benchmarks.
\newblock In {\em Proceedings of Annual Symposium on Combinatorial Search
  (SOCS)}, 2019.

\bibitem[\protect\citeauthoryear{{\v{S}}vestka and
  Overmars}{1998}]{vsvestka1998coordinated}
Petr {\v{S}}vestka and Mark~H Overmars.
\newblock Coordinated path planning for multiple robots.
\newblock {\em Robotics and autonomous systems}, 1998.

\bibitem[\protect\citeauthoryear{Van Den~Berg and
  Overmars}{2005}]{van2005prioritized}
Jur~P Van Den~Berg and Mark~H Overmars.
\newblock Prioritized motion planning for multiple robots.
\newblock In {\em Proceedings of IEEE/RSJ International Conference on
  Intelligent Robots and Systems (IROS)}, 2005.

\bibitem[\protect\citeauthoryear{Wagner and
  Choset}{2015}]{wagner2015subdimensional}
Glenn Wagner and Howie Choset.
\newblock Subdimensional expansion for multirobot path planning.
\newblock {\em Artificial Intelligence (AIJ)}, 2015.

\bibitem[\protect\citeauthoryear{Wagner \bgroup \em et al.\egroup
  }{2012}]{wagner2012probabilistic}
Glenn Wagner, Minsu Kang, and Howie Choset.
\newblock Probabilistic path planning for multiple robots with subdimensional
  expansion.
\newblock In {\em Proceedings of IEEE International Conference on Robotics and
  Automation (ICRA)}, 2012.

\bibitem[\protect\citeauthoryear{Wurman \bgroup \em et al.\egroup
  }{2008}]{wurman2008coordinating}
Peter~R Wurman, Raffaello D'Andrea, and Mick Mountz.
\newblock Coordinating hundreds of cooperative, autonomous vehicles in
  warehouses.
\newblock {\em AI magazine}, 2008.

\end{thebibliography}
\appendix
\section*{Appendix}

\section{Hyperparameter Adjustment}
\label{sec:hypra}

\subsection{Main Results}
Each algorithm in \cref{sec:result} has hyperparameters, e.g., the maximum distance to connect two vertices in the PRM-based two-phase planning methods (PP and CBS).
For each scenario and each algorithm, we adjusted hyperparameters to maximize the number of successful instances within \SI{30}{\second} of 50 instances, among randomly chosen 50 pairs of parameters.
Tie-break was based on average runtime.
The used instances were generated following the same parameters of the experiment but differed in random seeds.
The used parameters are included in the configuration files of the code.

\subsection{Scalability Test}
\label{sec:appendix:scalability}
In \cref{sec:scalability}, we adjusted the parameters manually and applied them to all instances regardless of $|A|$.
SSSP used $m=10$, $\theta_i = 0.05$, and $\epsilon = 0.2$ (in the \steer function).
PP was run on PRMs with $500$ samples and the maximum distance to connect two vertices was $0.1$.
These PRM's values were adjusted such that roadmaps sufficiently cover the entire workspace while being not too dense;
otherwise, the roadmap construction itself takes too much time and the search space will become too huge.
\Cref{fig:roadmap-example-scalability} shows two roadmaps created by PP (PRM) and SSSP.
We informally observed that these PP parameters were sensitive to obtain consistently good results.

{
  \setlength{\tabcolsep}{0pt}
  \newcommand{\figcol}[2]{
    \begin{minipage}{0.48\linewidth}
      \centering
      {\small #2}\\
      \includegraphics[width=1\linewidth]{fig/raw/#1}
    \end{minipage}
  }

  \begin{figure}[th!]
    \centering
    \scalebox{0.9}{
    \begin{tabular}{cc}
      \figcol{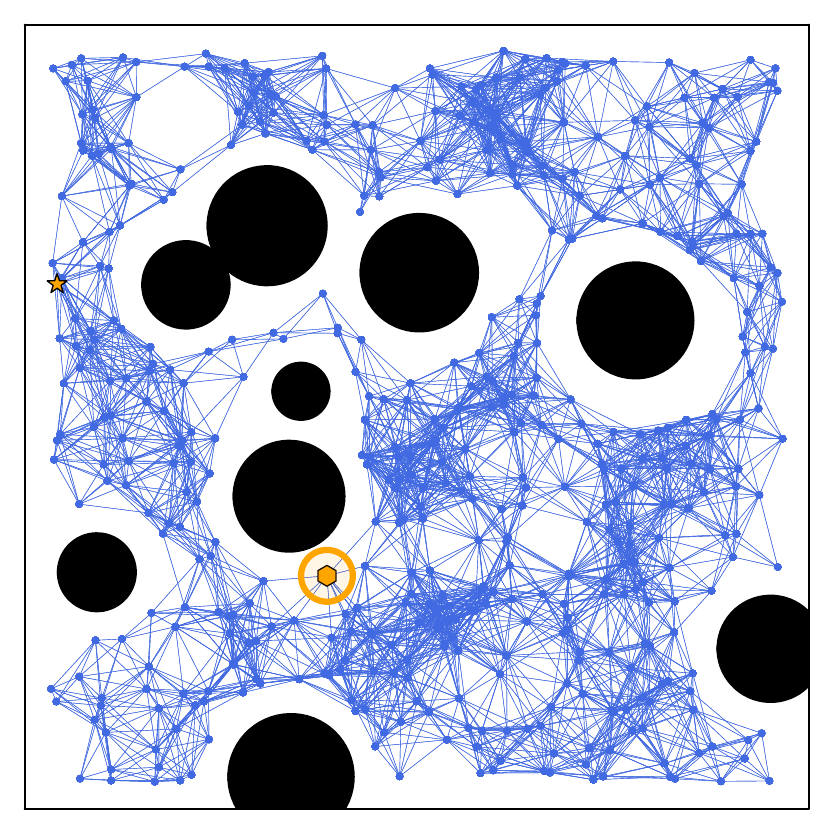}{PP (PRM)}
      &
        \figcol{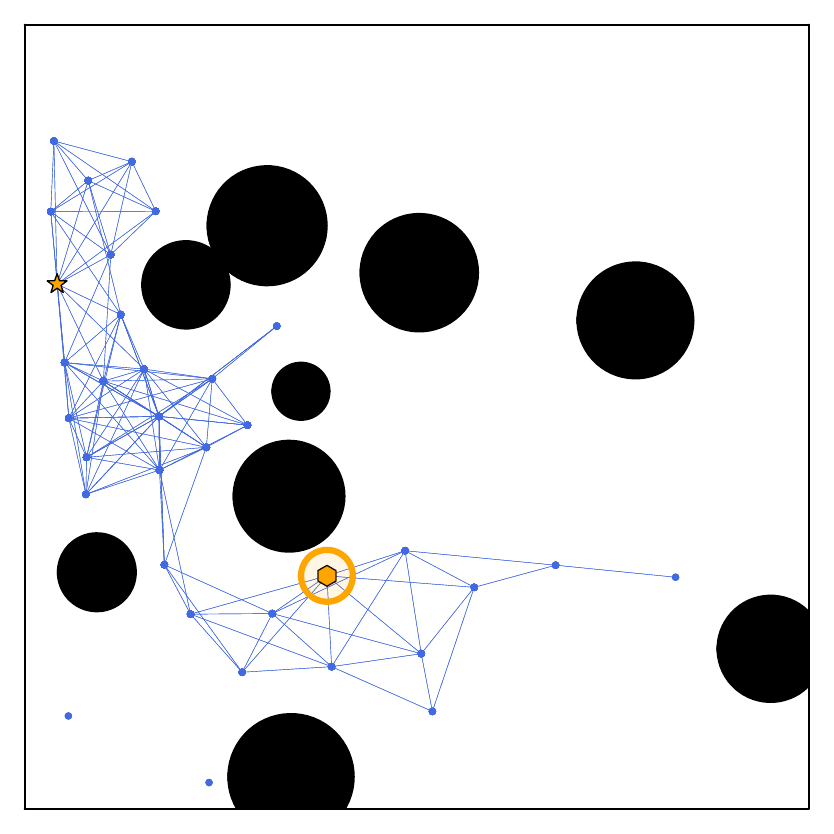}{SSSP}
    \end{tabular}
  }
    \caption{
      Constructed roadmaps for the scalability test.
    }
    \label{fig:roadmap-example-scalability}
  \end{figure}
}

\paragraph{Remark for Experimental Environment.}
We note that the scalability test was heavily affected by experimental environments.
We informally confirmed that both PP and SSSP could be faster ($\geq$x2) with another environment.
Nevertheless, SSSP generally worked better than the baseline.
For instance, SSSP solved instances with $50$ agents ($\leq$\SI{5}{\minute}) while PP failed the same instances.

\section{Effect of Steering}
We additionally evaluated SSSP with different $\lambda$ (probability of vanilla random sampling) in \emph{point2d} with $20$ robots, which were the same instances as \cref{sec:scalability}.
To illustrate the steering effect clearly, the hyperparameters of SSSP were set as $m=5$, $\theta_i = 0.02$, and $\epsilon = 0.05$.
\Cref{fig:result-steering} shows the results, demonstrating the decrease in success rate with a higher rate of vanilla random sampling.

{
  \begin{figure}[th!]
    \centering
    \includegraphics[width=0.8\linewidth]{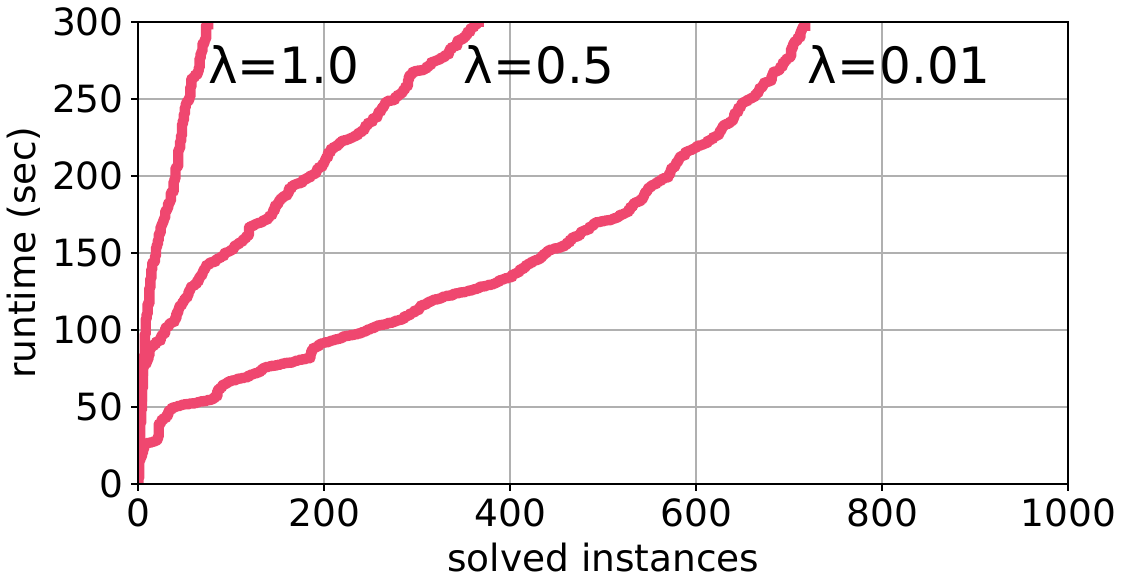}
    \caption{
      Effect of steering.
      For each instance of \scenario{point2d} among 100 instances, where $|A|=20$, SSSP was run 10 times with different random seeds.
    }
    \label{fig:result-steering}
  \end{figure}
}

As seen, vanilla random sampling is convenient from the theoretical side.
In practice, however, $\lambda$ should be set small to develop roadmaps in important regions for the multi-agent search.

\section{Solution Quality of Ablation Study}
\Cref{table:result-solution-quality-ablation} shows total traveling time of the ablation study in \cref{sec:ablation}.
From the table, we observe that initial roadmap construction contributes to improving solution quality.
This is because, with initial roadmaps, each roadmap is expanded mainly in neighboring regions of a valid path from start to goal included in the initial one.
On the other hand, such a ``guide'' does not exist without initial roadmaps, resulting in non-efficient trajectories.

{
  \renewcommand{\arraystretch}{1.5}
  \setlength{\tabcolsep}{0.5pt}
  \newcommand{\w}[1]{\textbf{#1}}
  \newcommand{\ci}[1]{\tiny{(#1)}}
  \newcommand{\head}[1]{
    {\renewcommand{\arraystretch}{0.8}\begin{tabular}{c}#1\end{tabular}}
  }
  \newcommand{\myblock}[1]{\renewcommand{\arraystretch}{0.5}\begin{tabular}{c}#1\end{tabular}}
  \newcommand{\rowhead}[2]{
    \renewcommand{\arraystretch}{0.5}
    \begin{tabular}{r}
      \scenario{#1}\\{\tiny #2}
    \end{tabular}
  }
  \begin{table}[th!]
    \centering
    \scriptsize
    \begin{tabular}{rcccccccc}
      \toprule
      & SSSP
      & \head{random\\score}
      & \head{no init\\roadmap}
      & \head{no vertex\\expansion}
      & \head{no \dist\\check}
      & \head{$\lambda$$=$$1$}
      & \head{on\\PRM}
      \\\midrule
      \rowhead{point2d}{54/100}
      & \myblock{\w{1.96}\\\ci{1.75, 2.15}}
      & \myblock{7.49\\\ci{5.44, 9.28}}
      & \myblock{8.75\\\ci{4.78, 11.84}}
      & \myblock{\w{1.89}\\\ci{1.63, 2.08}}
      & \myblock{\w{1.70}\\\ci{1.58, 1.81}}
      & \myblock{\w{2.02}\\\ci{1.80, 2.21}}
      & \myblock{\w{1.78}\\\ci{1.58, 1.95}}
      \smallskip\\
      \rowhead{arm22}{52/100}
      & \myblock{\w{1.76}\\\ci{1.58, 1.91}}
      & \myblock{\NA}
      & \myblock{7.38\\\ci{5.71, 8.95}}
      & \myblock{\w{1.87}\\\ci{1.64, 2.06}}
      & \myblock{\w{1.69}\\\ci{1.55, 1.82}}
      & \myblock{\w{1.74}\\\ci{1.58, 1.88}}
      & \myblock{\w{1.93}\\\ci{1.72, 2.12}}
      \smallskip\\
      \rowhead{snake2d}{63/100}
      & \myblock{3.35\\\ci{3.08, 3.62}}
      & \myblock{\NA}
      & \myblock{50.11\\\ci{40.72, 58.44}}
      & \myblock{3.78\\\ci{3.41, 4.13}}
      & \myblock{\NA}
      & \myblock{\w{2.77}\\\ci{2.56, 2.97}}
      & \myblock{3.64\\\ci{3.35, 3.93}}
      \\\bottomrule
    \end{tabular}
    \caption{
      Total traveling time of ablation study.
      See also the caption of \cref{table:result-solution-quality}.
    }
    \label{table:result-solution-quality-ablation}
  \end{table}
}

\end{document}